\documentclass[runningheads]{llncs}
\usepackage{graphicx}
\usepackage{cite}
\usepackage[pagebackref=false,breaklinks=true,letterpaper=true,colorlinks,urlcolor=blue,citecolor=blue,linkcolor=blue,bookmarks=false]{hyperref}
\usepackage{subcaption}
\usepackage{amsmath,amssymb}
\usepackage{booktabs}
\usepackage{hyperref}

\newcommand{\ra}[1]{\renewcommand{\arraystretch}{#1}}

\newcommand{\etal}{\textit{et~al}\mbox{.}}

\newcommand{\ie}{i.e.,\ }
\newcommand{\bbR}{{\mathbb{R}}}

\newlength{\threeimg}
\newlength{\fourimg}
\newlength{\fiveimg}
\newlength{\siximg}

\newlength\paramargin
\newlength\figmargin
\newlength\secmargin
\newlength\figcapmargin

\begin{document}

\title{Deep Semantic Matching with Foreground Detection and Cycle-Consistency} 
\titlerunning{Deep Semantic Matching with Foreground Detection and Cycle-Consistency} 


\author{
Yun-Chun Chen\inst{1,2} \and 
Po-Hsiang Huang\inst{2} \and 
Li-Yu Yu\inst{2} \and
Jia-Bin Huang\inst{3} \and 
Ming-Hsuan Yang\inst{4,5} \and 
Yen-Yu Lin\inst{1}
}

\authorrunning{Y.-C. Chen, P.-H. Huang, L.-Y. Yu, J.-B. Huang, M.-H. Yang, and Y.-Y. Lin} 

\institute{
$^1$Academia Sinica \; $^2$National Taiwan University \; $^3$Virginia Tech \; \\
$^4$University of California, Merced \; $^5$Google Cloud
}

\maketitle

\begin{abstract}
Establishing dense semantic correspondences between object instances remains a challenging problem due to background clutter, significant scale and pose differences, and large intra-class variations.
In this paper, we address weakly supervised semantic matching based on a deep network where only image pairs without manual keypoint correspondence annotations are provided.
To facilitate network training with this weaker form of supervision, we 1) explicitly estimate the foreground regions to suppress the effect of background clutter and 2) develop cycle-consistent losses to enforce the predicted transformations across multiple images to be geometrically plausible and consistent.
We train the proposed model using the PF-PASCAL dataset and evaluate the performance on the PF-PASCAL, PF-WILLOW, and TSS datasets.
Extensive experimental results show that the proposed approach performs favorably against the state-of-the-art methods.
%
\end{abstract}

\section{Introduction}

Dense correspondence matching is an important and active research topic in computer vision. Optical flow estimation~\cite{horn1981determining,lucas1981iterative} and stereo matching~\cite{StereoMatching,StereoMatch} aim to estimate per-pixel correspondences to match across images depicting the same scene or object instance. While correspondence estimation has been extensively studied, there has been a growing trend to extend the idea of matching the same objects across images to matching images covering \emph{different instances} of an object category. This progress not only attracts substantial attention but also facilitates many real-world applications ranging from object recognition~\cite{SIFTFlow}, object co-segmentation~\cite{chen2015co,Taniai,hsuco}, to 3D reconstruction~\cite{mustafa}. However, due to the presence of background clutter, ambiguity induced by large intra-class variations, and the limited scalability of obtaining large-scale datasets with manually annotated correspondences, semantic matching remains challenging.

Conventional methods for semantic matching rely on hand-crafted descriptors such as SIFT~\cite{SIFTFlow} or HOG~\cite{HoG} as well as an effective geometric regularizer. However, these hand-crafted descriptors cannot be adapted to the given visual domains, leading to  a sub-optimal performance of semantic matching. Driven by the recent success of convolutional neural networks (CNNs), several learning-based approaches have been proposed for tackling the problem of semantic matching~\cite{UCN,SCNet,FCSS,CNNGeo,End-to-end}. While promising results have been shown, these approaches still suffer from the following limitations. The methods in~\cite{UCN,SCNet,FCSS,CNNGeo} require a vast amount of supervised data for training the network. Collecting large-scale and diverse data, however, is expensive and labor-intensive. While weakly supervised methods such as~\cite{End-to-end} have been recently proposed to relax the issue, these approaches implicitly match the background features from both images to be similar. Thus, they often suffer from the unfavorable effect of background clutter.

\setlength{\threeimg}{0.185\linewidth}
\begin{figure}[t]
  \centering
  \begin{subfigure}[b]{\threeimg}
    \includegraphics[width=0.98\linewidth]{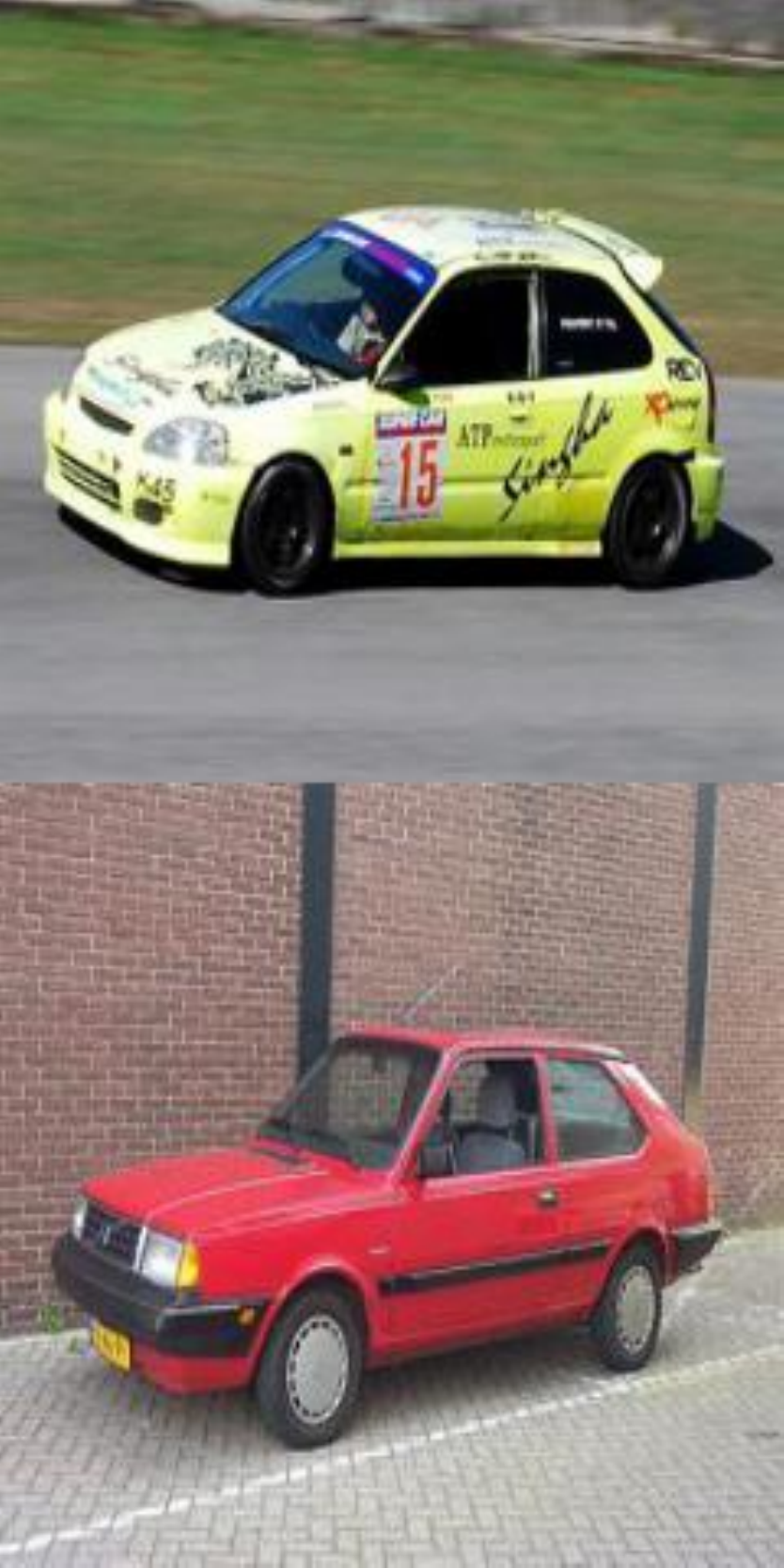}
    \caption*{Input}
    \label{fig:input}
  \end{subfigure}
  \begin{subfigure}[b]{2.0\threeimg}
    \includegraphics[width=0.49\linewidth]{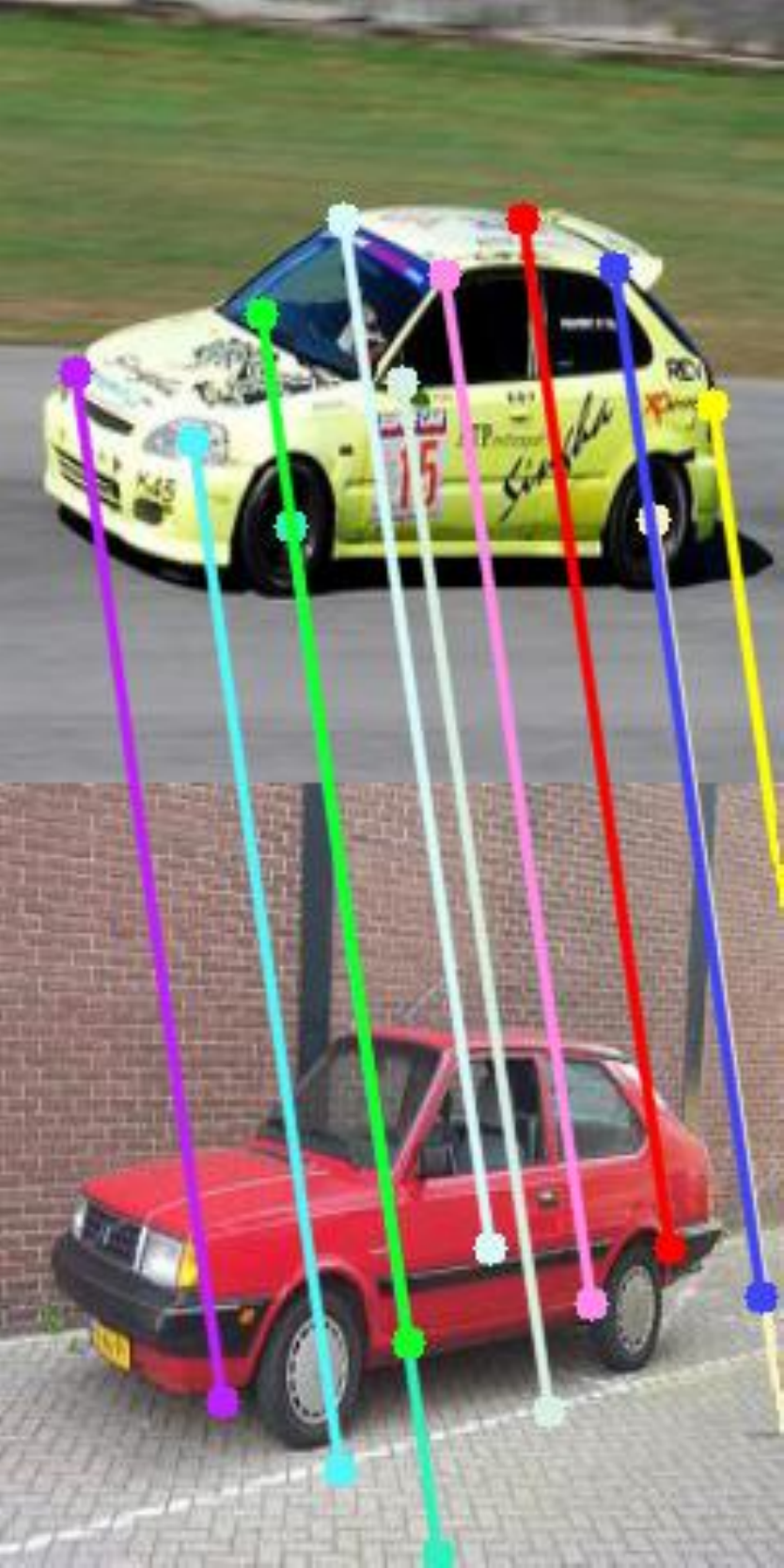}%
    \hfill
    \includegraphics[width=0.49\linewidth]{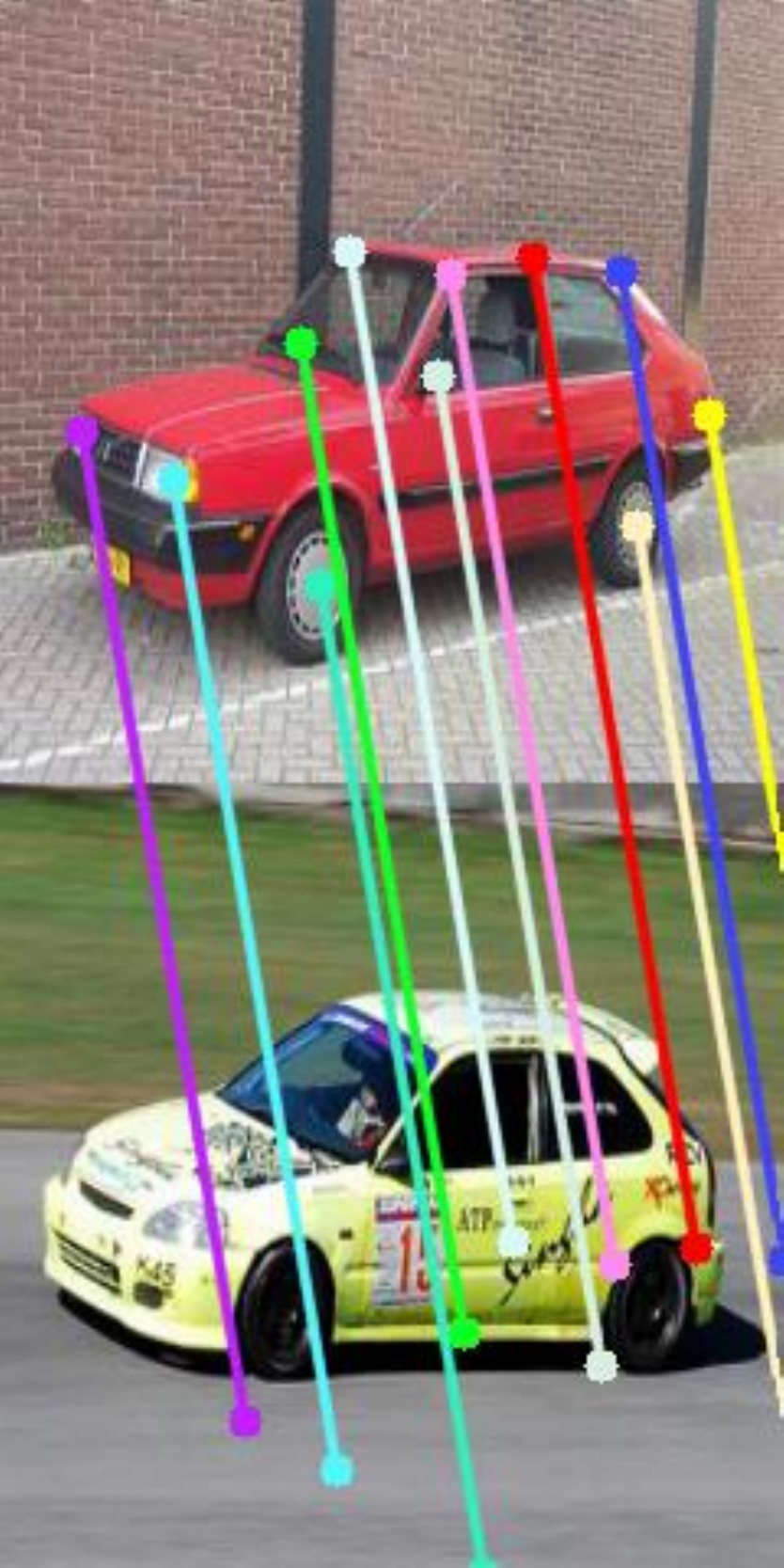}
    \centering\caption*{Rocco~\etal~\cite{End-to-end}}
    \label{fig:baseline}
  \end{subfigure}
  \begin{subfigure}[b]{2.0\threeimg}
    \includegraphics[width=0.49\linewidth]{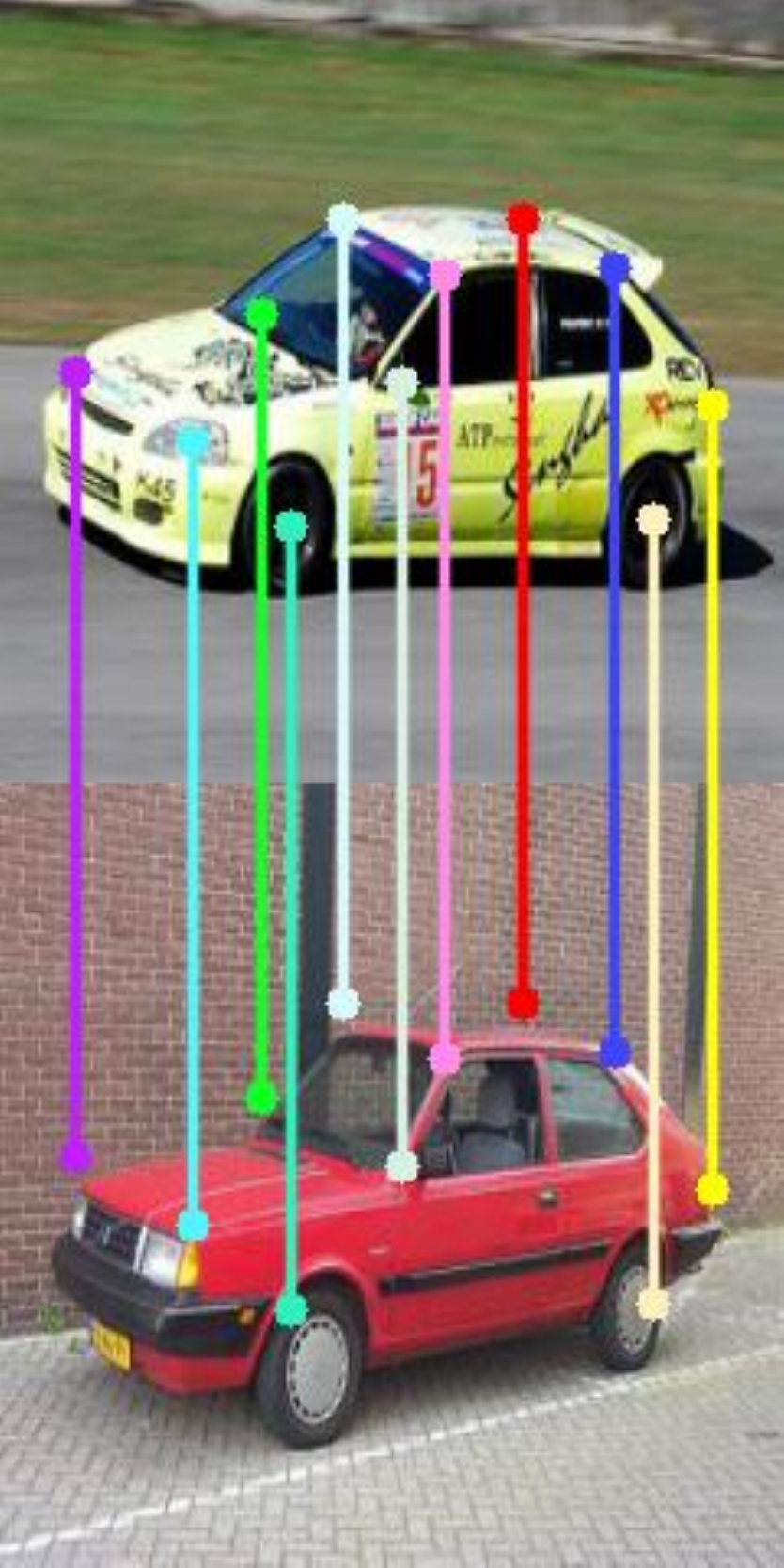}%
    \hfill
    \includegraphics[width=0.49\linewidth]{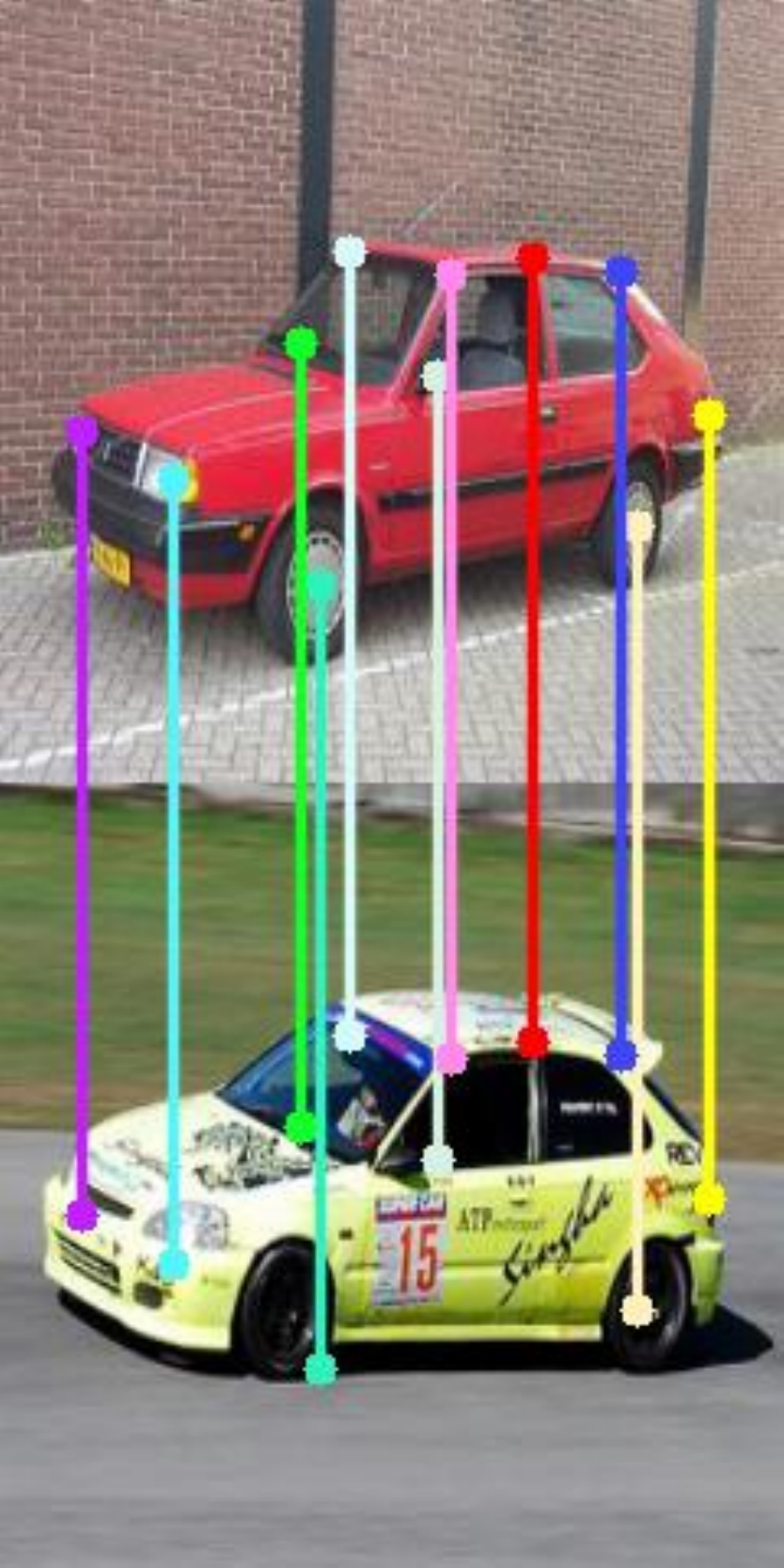}
    \centering\caption*{Ours}
    \label{fig:ours-1}
  \end{subfigure}
  \caption{\textbf{Comparisons with the state-of-the-art semantic matching algorithm~\cite{End-to-end}.} Existing semantic matching methods often suffer from background clutter and may produce inconsistent matching results when swapping the source/target image. Through integrating foreground detection and cycle-consistent checking into semantic matching, our method produces more accurate and consistent matching results in both directions.}
  \label{fig:teaser}
\end{figure}

In this paper, we address these challenges by performing foreground detection and enforcing cycle consistency constraints in semantic matching. To suppress the negative impacts caused by background clutter, we develop a foreground detection module that allows the model to exclude background regions and focus on matching the detected foreground regions. Therefore, the effect of background clutter can be alleviated. To address the matching difficulties caused by complex appearance and large intra-class variations, we focus on filtering out geometrically inconsistent correspondences. Our key insight is that correct correspondence should be \emph{cycle-consistent} meaning that when matching a particular point from one image to the other and then performing reverse matching, we should arrive at the same spot. To exploit this property, we introduce a cycle-consistency loss that provides additional supervisory signals for network training. We further extend this idea to explore transitivity consistency across multiple images. We build upon the model by Rocco~\etal~\cite{End-to-end} for a weakly-supervised and end-to-end trainable network and evaluate the effectiveness of the proposed approach on three benchmarks. Experimental results demonstrate that our approach improves the baseline model~\cite{End-to-end}, as shown in Fig.~\ref{fig:teaser}, and performs favorably against the state-of-the-art methods.

Our contributions are summarized as follows. First, we present a weakly-supervised learning framework that integrates foreground detection into semantic matching. With a module for explicit foreground detection, the proposed network suppresses the unfavorable effect of background clutter. Second, our model implicitly tackles the ambiguity induced by a vast matching space via inferring bi-directional geometric transformations during matching. With these transformations, we explicitly enforce the inferred geometric transformations to be cycle-consistent by introducing the \textit{forward-backward consistency loss}. In addition, we explore the property of transitivity consistency and introduce the \textit{transitivity consistency loss} to further enhance the matching performance. We train our network with the image pairs of the PF-PASCAL dataset~\cite{ProposalFlow}. We then evaluate the proposed model on several benchmark datasets for semantic matching, including the PF-PASCAL~\cite{ProposalFlow}, PF-WILLOW~\cite{ProposalFlow}, and TSS~\cite{Taniai} datasets. Extensive comparisons with existing semantic matching algorithms demonstrate that the proposed approach achieves the state-of-the-art performance.

\section{Related Work}\label{sec:RelatedWork}

Semantic matching has been extensively studied in the literature. Here, we review several related topics.

{\flushleft {\bf Semantic matching.}}
Conventional methods for semantic matching~\cite{hu2016progressive,hu2015matching,hsu2015robust} leverage hand-crafted descriptors such as SIFT~\cite{SIFTFlow} or HOG~\cite{HoG} along with geometric matching models. These methods find keypoint correspondences across images through energy minimization. The SIFT Flow~\cite{SIFTFlow} method aligns two images with SIFT features~\cite{SIFTFlow} using a similar formulation as an optical flow algorithm. Kim~\etal~\cite{Deformable} compute dense correspondence efficiently using the deformable spatial pyramid. Ham~\etal~\cite{ProposalFlow} use the object proposals as the matching primitives and leverage the HOG descriptor to establish semantic correspondence. With the use of object proposals, the Proposal Flow method is robust to scaling and background clutter. Taniai~\etal~\cite{Taniai} propose a hierarchical Markov random field model to jointly perform object co-segmentation and dense correspondence. However, the aforementioned methods adopt hand-crafted descriptors, which are pre-defined and not optimized for the given images. 

{\flushleft {\bf Semantic matching via deep learning.}}
Convolutional neural networks have been successfully applied to semantic matching. Choy~\etal~\cite{UCN} propose the universal correspondence network (UCN) and a correspondence contrastive loss for network training. The UCN method adopts a convolutional spatial transformer for feature transformations, making their method robust to scaling and rotations. Kim~\etal~\cite{FCSS} propose the fully convolutional self-similarity (FCSS) descriptor and integrate the descriptor into the Proposal Flow framework~\cite{ProposalFlow} for image matching. The SCNet~\cite{SCNet} method learns a geometrically plausible model for semantic correspondence by incorporating geometric consistency constraints into its loss function. While the methods in~\cite{FCSS,SCNet} employ trainable descriptors for semantic correspondence, the feature matching is learned at the \emph{object-proposal} level. Consequently, these methods are not end-to-end trainable since a fusion step is required to produce the final results. Rocco~\etal~\cite{CNNGeo} present an end-to-end trainable CNN architecture for estimating parametric geometric transformations. While these methods~\cite{ProposalFlow,UCN,FCSS,SCNet,CNNGeo} perform better than those based on hand-crafted features, the dependence on supervised training data with manually labeled keypoint correspondences limits the applicability.

Several recent CNN-based methods~\cite{AnchorNet,End-to-end,WarpNet,chen2019show} have carried out weakly supervised semantic correspondence. The AnchorNet~\cite{AnchorNet} learns a set of filters whose response is geometrically consistent across different object instances. The AnchorNet model, however, is not end-to-end trainable due to the use of the hand-engineered alignment model. The WarpNet~\cite{WarpNet} learns fine-grained image matching with small-scale and pose variations via aligning objects across images through known deformation. Inspired by the inlier scoring procedure of RANSAC, Rocco~\etal~\cite{End-to-end} propose an end-to-end trainable alignment network which computes dense semantic correspondence while aligning two images.

Our proposed method differs from these methods~\cite{AnchorNet,End-to-end} in two aspects. First, our approach further takes into account foreground detection. Our network learns feature embedding to enhance inter-image foreground similarity while alleviating the unfavorable effects caused by complex background. Second, our model simultaneously infers bi-directional transformations. We explicitly enforce cycle-consistent constraints on the predicted transformations, resulting in more accurate and consistent matching results.

{\flushleft {\bf Cycle consistency.}}
Exploiting cycle consistency property to regularize learning has been extensively studied. In the context of motion analysis, computing bi-directional optical flow has been shown to be useful to reason about occlusion for learning optical flow~\cite{UnFlow,DFNet} and enforcing temporal consistency~\cite{huang2016temporally,lai2018learning}. In the context of image-to-image translation, enforcing cycle consistency enables learning mapping between domains with unpaired data~\cite{CycleGan,DRIT}. In the context of unsupervised domain adaptation, exploiting cycle consistency constraints allows the model to produce consistent task predictions across domains~\cite{chen2019crdoco}. In the context of visual recognition~\cite{chen2019learning,li2019recover}, enforcing consistency constraints allows the model to be more robust to resolution variations~\cite{li2020cross}. Several methods exploit the idea of cycle consistency for semantic matching. Zhou~\etal~\cite{Multi-match} tackle the problem of matching multiple images by jointly optimizing feature matching and enforcing cycle consistency. The FlowWeb~\cite{FlowWeb} method learns image alignment by establishing globally-consistent dense correspondences with cycle consistency constraints. However, these methods~\cite{Multi-match,FlowWeb} employ hand-engineered descriptors which cannot adapt to an arbitrary object category given for matching. Zhou~\etal~\cite{3D-cycle} establish dense correspondences by using an additional 3D CAD model to form a cross-instance loop between synthetic data and real images. However, the cycle consistency loss in~\cite{3D-cycle} requires four images at a time. In contrast, we develop two loss functions to enforce cycle consistency and do not need additional data to guide the training. Experimental results demonstrate that by exploiting cycle consistency constraints, the proposed method produces consistent matching results and improves the performance.

\section{Proposed Algorithm} \label{sec:ProposedMethod}

In this section, we first provide an overview of our approach. We then describe each loss in our objective function in detail and the implementation details.

\begin{figure}[t]
  \centering
  \includegraphics[width=\linewidth]{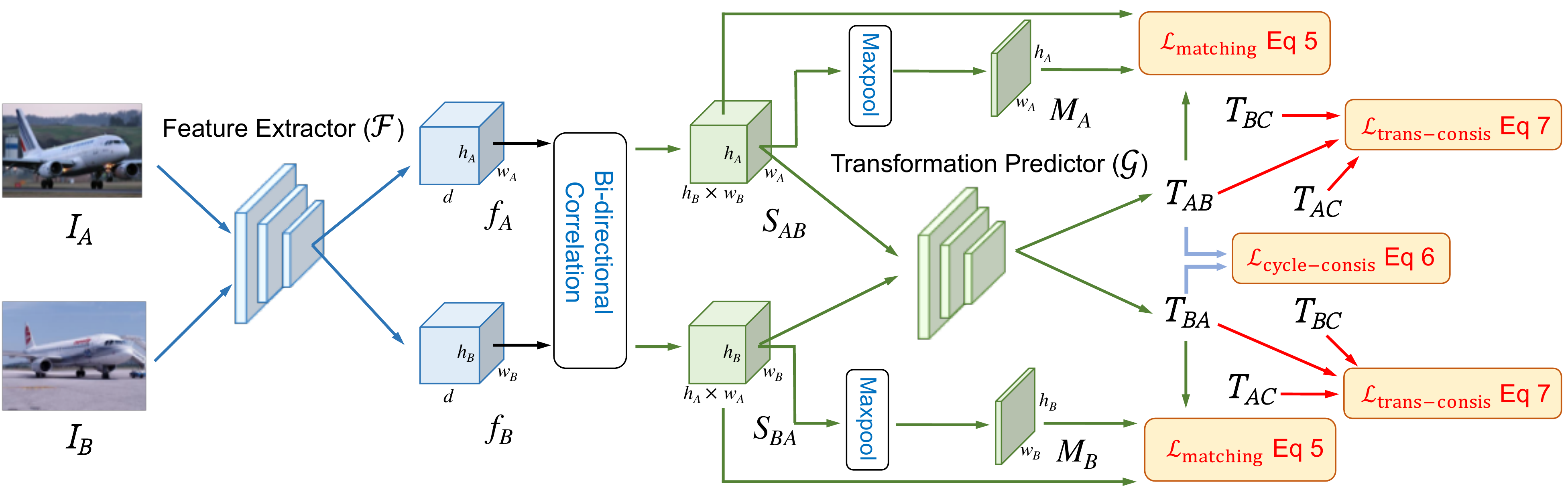}
  \caption{\textbf{Semantic matching network.} Our model is composed of two CNN modules, including a feature extractor $\mathcal{F}$ for extracting features and a transformation predictor $\mathcal{G}$ for estimating the geometric transformations between a given image pair. We train the model with three loss functions, including the foreground-guided matching loss $\mathcal{L}_\mathrm{matching}$, the forward-backward consistency loss $\mathcal{L}_\mathrm{cycle-consis}$, and the transitivity consistency loss $\mathcal{L}_\mathrm{trans-consis}$ (if given 3 input images).}
  \label{fig:Model}
\end{figure}

\subsection{Framework overview}
Let $\mathcal{D} = \{I_i\}_{i=1}^N$ denote a set of images consists of instances of the same object category, where $I_i$ is the $i^{th}$ image and $N$ is the number of images. Our goal is to learn a CNN-based model that can estimate the keypoint correspondences between each image pair $(I_A, I_B)$ in $\mathcal{D}$ \emph{without knowing the object class in advance}. Our formulation for semantic matching is \emph{weakly-supervised} since training our model requires only weak image-level supervision in the form of training image pairs containing common objects. No ground truth keypoint correspondences are used.

To accomplish this task, we present an end-to-end trainable network which is composed of two modules: 1) the feature extractor $\mathcal{F}$ and 2) the transformation predictor $\mathcal{G}$. The feature extractor $\mathcal{F}$ extracts features for each image in a given image pair. The transformation predictor $\mathcal{G}$ predicts the transformation that warps an image so that the warped image can better align the other image.

As shown in Fig.~\ref{fig:Model}, the proposed network architecture takes an image pair as input. For a given image pair $(I_A, I_B)$, we use the feature extractor $\mathcal{F}$ to extract their feature maps $f_A$ and $f_B$, respectively. We then compute correlation from $f_A$ to $f_B$ to generate the correlation map $S_{AB}$. The other correlation map $S_{BA}$ is symmetrically obtained. The transformation predictor $\mathcal{G}$ then estimates the geometric transformation $T_{AB}$ which warps $I_A$ so that the warped image $\tilde{I}_A$ can align $I_B$. In the following section, we describe our objective function used to optimize the feature extractor $\mathcal{F}$ and the transformation predictor $\mathcal{G}$. After optimizing the objective function, the matching between an image pair $(I_A,I_B)$ can be performed via the predicted transformation $T_{AB}$ or $T_{BA}$.

\subsection{Objective function}
The overall training objective consists of three loss functions. First, the foreground-guided matching loss $\mathcal{L}_\mathrm{matching}$ minimizes the distance between the corresponding features based on the estimated geometric transformations. Unlike existing semantic matching methods~\cite{CNNGeo,End-to-end}, our model predicts foreground masks to suppress the effect of background clutter by excluding background matching. Second, the forward-backward consistency loss $\mathcal{L}_\mathrm{cycle-consis}$ and the transitivity consistency loss $\mathcal{L}_\mathrm{trans-consis}$ enforce the predicted transformations across multiple images to be geometrically plausible and consistent. Both losses regularize the network training. Specifically, our training objective is
\begin{equation}
  \mathcal{L} = \mathcal{L}_\mathrm{matching} + \lambda_C\cdot\mathcal{L}_\mathrm{cycle-consis} + \lambda_T\cdot\mathcal{L}_\mathrm{trans-consis},
  \label{eq:FullObj}
\end{equation}
where $\lambda_C$ and $\lambda_T$ are hyper-parameters used to control the relative importance of the respective loss functions. Below we outline the details of each loss function.

\subsection{Foreground-guided matching loss}\label{loss:Mask}
To reduce the effect of background clutter and enforce only foreground regions to be similar, our model minimizes the foreground-guided matching loss $\mathcal{L}_\mathrm{matching}$. Given an image pair $(I_A,I_B)$, the feature extractor $\mathcal{F}$ extracts their respective feature maps $f_A \in \mathbb{R}^{h_A\times w_A\times d}$ and $f_B \in \mathbb{R}^{h_B\times w_B\times d}$, where $d$ is the number of channels. We correlate $f_A$ with $f_B$ to generate the correlation map $S_{AB} \in \mathbb{R}^{h_A\times w_A\times h_B\times w_B}$. Each element $S_{AB}(i,j,s,t) = S_{AB}(\mathbf{p},\mathbf{q})$ records the normalized inner product between the feature vectors stored at two spatial locations $\mathbf{p} = [i, j]^{\top}$ in $f_A$ and $\mathbf{q} = [s, t]^{\top}$ in $f_B$. The other correlation map $S_{BA} \in \mathbb{R}^{h_B\times w_B\times h_A\times w_A}$ can be computed symmetrically. The correlation map $S_{AB}$ is reshaped to a three-dimensional tensor with dimensions $h_A$, $w_A$, and $(h_B\times w_B)$, \ie $S_{AB} \in \bbR^{h_A\times w_A\times(h_B\times w_B)}$. As such, the reshaped correlation map $S_{AB}$ can be interpreted as a dense $h_A\times w_A$ grid with $(h_B\times w_B)$-dimensional local features. We apply the reshaping operation to $S_{BA}$ as well. With the reshaped $S_{AB}$, we use the transformation predictor $\mathcal{G}$~\cite{CNNGeo} to estimate a geometric transformation $T_{AB}$ which warps $I_A$ to $\tilde{I}_A$ so that $\tilde{I}_A$ aligns well to $I_B$.

Since the correlation map $S_{AB}(\mathbf{p}, \mathbf{q})$ records the normalized inner product between two feature vectors located at $\mathbf{p}$ in $f_A$ and $\mathbf{q}$ in $f_B$. Our model estimates the foreground mask $M_A \in \mathbb{R}^{h_A \times w_A}$ by 
\begin{equation}
  M_A (\mathbf{p}) = \underset{\mathbf{q}}{\max}(S_{AB}(\mathbf{p},\mathbf{q})).
  \label{eq:Mask}
\end{equation}

Note that both the correlation maps $S_{AB}$ and $S_{BA}$ are compiled through a rectified linear unit (ReLU) to eliminate negative matching values in advance. Therefore, the value of the estimated foreground masks at each pixel will be bounded between $0$ and $1$. Intuitively, the mask $M_A(\mathbf{p})$ has a low value (i.e., location $\mathbf{p}$ is likely to belong to background) if none of the feature vectors in $f_B$ matches well with $f_A(\mathbf{p})$. The mask $M_B$ can be obtained following a similar procedure.

With the estimated geometric transformation $T_{AB}$, we can identify and filter out geometrically inconsistent correspondences. Consider a correspondence with endpoints $(\mathbf{p} \in \mathcal{P}_A, \mathbf{q} \in \mathcal{P}_B)$, where $\mathcal{P}_A$ and $\mathcal{P}_B$ are the sets of all spatial coordinates of $f_A$ and $f_B$, respectively. The distance $\|T_{AB}(\mathbf{p}) - \mathbf{q}\|$ represents the projection error of this correspondence with respect to transformation $T_{AB}$. Following Rocco~\etal~\cite{End-to-end}, we introduce a correspondence mask $m_A$ to determine if the correspondences are geometrically consistent with transformation $T_{AB}$. Specifically, $m_A$ is of the form
\begin{equation}
  m_{A}{(\mathbf{p},\mathbf{q})} = 
  \begin{cases}
    1, & \text{if $\|T_{AB}(\mathbf{p}) - \mathbf{q}\| \leq \varphi$},\\
    0, & \text{otherwise}.
  \end{cases}\mbox{, for $\mathbf{p}\in\mathcal{P}_A$ and $\mathbf{q}\in\mathcal{P}_B$,}
\label{eq:s}
\end{equation}
where $\varphi = 1$ is the number of pixels.

Given the geometric transformation $T_{AB}$ and the correspondence mask $m_A$, we compute matching score of each spatial location $\mathbf{p} \in \mathcal{P}_A$ as
\begin{equation}
  s_{A}{(\mathbf{p})} = \sum_{\mathbf{q}\in\mathcal{P}_{B}}^{}m_{A}{(\mathbf{p},\mathbf{q})}\cdot S_{AB}{(\mathbf{p},\mathbf{q})}.
  \label{eq:Score}
\end{equation}

To suppress the effect of background clutter, we incorporate the estimated foreground masks to focus on matching the detected foreground regions. We define the foreground-guided matching loss $\mathcal{L}_\mathrm{matching}$ as
\begin{equation}
  \mathcal{L}_\mathrm{matching}(I_A,I_B;\mathcal{F},\mathcal{G}) = - \bigg(\sum_{\mathbf{p}\in\mathcal{P}_{A}}^{}s_{A}{(\mathbf{p})}\cdot M_{A}{(\mathbf{p})} + \sum_{\mathbf{q}\in\mathcal{P}_{B}}^{}s_{B}{(\mathbf{q})}\cdot M_{B}{(\mathbf{q})}\bigg).
  \label{eq:MaskLoss}
\end{equation}

Note that the negative sign in (\ref{eq:MaskLoss}) is used in the objective function, since maximizing the matching score corresponds to minimizing the foreground-guided matching loss $\mathcal{L}_\mathrm{matching}$.

\subsection{Cycle consistency}
For a pair of images $I_A$ and $I_B$, the transformation predictor $\mathcal{G}$ estimates a geometric transformation $T_{AB}$ which maps pixel coordinates from $I_A$ to $I_B$. However, the large capacity of the transformation predictor $\mathcal{G}$ often leads to a circumstance where various transformations can warp $I_A$ to $\tilde{I}_A$ such that $\tilde{I}_A$ aligns $I_B$ very well. This phenomenon implies that using the foreground-guided matching loss $\mathcal{L}_\mathrm{matching}$ alone is insufficient to reliably train the transformation predictor $\mathcal{G}$ in the weakly supervised setting since no ground truth correspondences are available to constrain the learning of predicting transformations. We address this issue by simultaneously estimating $T_{AB}$ and $T_{BA}$ and enforce the predicted transformations to be geometrically plausible and consistent across multiple images. As such, exploiting the cycle consistency constraint greatly reduces the feasible space of transformations and can serve as a regularization term in training the transformation predictor $\mathcal{G}$. To this end, we develop two loss functions where cycle-consistency checking is performed in conjunction with the proposed method such that the model is end-to-end trainable. The developed loss functions are described in the following.

\subsubsection{Forward-backward consistency loss.} 
Consider the correlation maps $S_{AB}$ and $S_{BA}$ generated from images $I_A$ and $I_B$. The forward consistency states that property $T_{BA}(T_{AB}(\mathbf{p}))\approx \mathbf{p}$ holds for any $\mathbf{p}\in\mathcal{P}_{A}$. By the same token, the backward consistency means $T_{AB}(T_{BA}(\mathbf{q}))\approx \mathbf{q}$ for any $\mathbf{q}\in\mathcal{P}_{B}$. The resultant forward-backward consistency loss $\mathcal{L}_\mathrm{cycle-consis}$ is then defined by
\begin{equation}
  \begin{aligned}
    \mathcal{L}_\mathrm{cycle-consis}(I_A,I_B;\mathcal{F},\mathcal{G}) &{} = \sum_{\mathbf{p}\in\mathcal{P}_{A}}^{} \|T_{BA}(T_{AB}(\mathbf{p}))- \mathbf{p}\| \\
    & + \sum_{\mathbf{q}\in\mathcal{P}_{B}} \|T_{AB}(T_{BA}(\mathbf{q})) - \mathbf{q}\|,
  \end{aligned}
  \label{eq:Cycle}
\end{equation}
where $\|T_{BA}(T_{AB}(\mathbf{p})) - \mathbf{p}\|$ is the reprojection error between coordinate $\mathbf{p}$ and the reprojected coordinate $T_{BA}(T_{AB}(\mathbf{\mathbf{p}}))$.

\subsubsection{Transitivity consistency loss.}
We further extend the forward-backward consistency between a pair of images to the transitivity consistency across multiple images. Considering the case of three images $I_A$, $I_B$, and $I_C$, we first estimate three geometric transformations $T_{AB}$, $T_{BC}$, and $T_{AC}$. Transitivity consistency in this case states that the coordinate transformation from $I_A$ to $I_C$ should be path invariant. That is, for any coordinate $\mathbf{p}\in\mathcal{P}_{A}$, the property, $T_{BC}(T_{AB}(\mathbf{p}))\approx T_{AC}(\mathbf{p})$, holds. We can thus introduce the transitivity consistency loss $\mathcal{L}_\mathrm{trans-consis}$ as
\begin{equation}
  \begin{aligned}
    \mathcal{L}_\mathrm{trans-consis}(I_A,I_B,I_C;\mathcal{F},\mathcal{G}) &{} = \sum_{\mathbf{p}\in\mathcal{P}_{A}}^{}\|T_{BC}(T_{AB}(\mathbf{p})) - T_{AC}(\mathbf{p})\| \\
    & + \sum_{\mathbf{q}\in\mathcal{P}_{B}}^{}\|T_{AC}(T_{BA}(\mathbf{q})) - T_{BC}(\mathbf{q})\|.
  \end{aligned}
  \label{eq:Trans}
\end{equation}

\subsection{Network selection and initialization}

We adopt the semantic matching network proposed by Rocco~\etal~\cite{End-to-end} as our feature extractor $\mathcal{F}$ due to its state-of-the-art performance for image alignment. The network employs the ResNet-101~\cite{ResNet} model. The extracted features are those generated by layer \texttt{conv4-23}. For the transformation predictor $\mathcal{G}$, we use the same architecture as that in~\cite{CNNGeo}. The transformation predictor $\mathcal{G}$ is a cascade of two modules predicting an affine transformation and a thin plate spline (TPS) transformation. Given an image pair, the model first estimates an affine transformation with 6 degrees of freedom to obtain a rough alignment. The model then performs a second-stage geometric estimation based on the roughly aligned image pair to predict TPS transformation for alignment refinement. Similar to Rocco~\etal~\cite{CNNGeo}, we use a uniform $3 \times 3$ grid of control points for TPS, which corresponds to $3 \times 3 \times 2 = 18$ degrees of freedom. We initialize the feature extractor $\mathcal{F}$ and the transformation predictor $\mathcal{G}$ from the parameters pre-trained in \cite{End-to-end} and fine-tune the feature extractor $\mathcal{F}$ and the transformation predictor $\mathcal{G}$ by using the proposed objective function. We note that there may exist degenerate solutions to the foreground-guided matching loss $\mathcal{L}_\mathrm{matching}$ since no annotated correspondences are used to guide the network training. In this work, we build our model upon Rocco~\etal~\cite{End-to-end}, which is pre-trained on a large-scale synthetic dataset. The pre-trained model provides good enough initialization for predicting the geometric transformations, reducing the chance of falling into degenerate solutions. In addition, the foreground-guided matching loss $\mathcal{L}_\mathrm{matching}$ and the cycle-consistency losses work jointly. The three adopted loss terms regularize the network training and avoid degenerate solutions.

\section{Experimental Results} \label{sec:Experiments}
Experiments are conducted in this section. Here, we first describe the implementation details and the experimental setting. We evaluate and compare the proposed approach with the state-of-the-art, following analyzing the relative contributions of individual components in the proposed model.

\subsection{Implementation details}
We implement our model using PyTorch. We use the training and validation image pairs from the PF-PASCAL dataset~\cite{ProposalFlow}. All images are resized to the resolution of $240 \times 240$. We perform data augmentation by horizontal flipping, random cropping the input images, and swapping the order of images in the image pair. We train our model using the ADAM optimizer~\cite{Adam} with an initial learning rate of $5 \times 10^{-8}$. For transitivity consistency loss, the input triplets are randomly selected within a mini-batch. We sample $10 \times 10 = 100$ spatial coordinates for computing the forward-backward consistency loss and the transitivity consistency loss. The training process takes about 2 hours on a single NVIDIA GeForce GTX 1080 GPU.

\subsection{Evaluation metric and datasets} 
We conduct the evaluation on the PF-PASCAL~\cite{ProposalFlow}, PF-WILLOW~\cite{ProposalFlow}, and TSS~\cite{Taniai} benchmark datasets.

\begin{table}[t]
  \scriptsize
  \ra{1.2}
  \begin{center}
  \caption{\textbf{Per-class PCK on the PF-PASCAL dataset with $\tau = 0.1$.}}
  \label{table:PF-PASCAL}
  \resizebox{\linewidth}{!}
  {
  \begin{tabular}{l|cccccccccccccccccccc|c}
  \toprule
  Method & aero & bike & bird & boat & bottle & bus & car & cat & chair & cow & d.table & dog & horse & moto & person & plant & sheep & sofa & train & tv & mean\\
  \midrule
  HOG+PF-LOM~\cite{ProposalFlow}  & 73.3 & 74.4 & 54.4 & 50.9 & 49.6 & 73.8 & 72.9 & 63.6 & 46.1 & 79.8 & 42.5 & 48.0 & 68.3 & 66.3 & 42.1 & 62.1 & 65.2 & 57.1 & 64.4 & 58.0 & 62.5\\
  UCN~\cite{UCN} & 64.8 & 58.7 & 42.8 & 59.6 & 47.0 & 42.2 & 61.0 & 45.6 & 49.9 & 52.0 & 48.5 & 49.5 & 53.2 & 72.7 & 53.0 & 41.4 & 83.3 & 49.0 & \textbf{73.0} & 66.0 & 55.6\\
  VGG-16+SCNet-A~\cite{SCNet} & 67.6 & 72.9 & 69.3 & 59.7 & 74.5 & 72.7 & 73.2 & 59.5 & 51.4 & 78.2 & 39.4 & 50.1 & 67.0 & 62.1 & \textbf{69.3} & 68.5 & 78.2 & 63.3 & 57.7 & 59.8 & 66.3\\
  VGG-16+SCNet-AG~\cite{SCNet} & 83.9 & 81.4 & 70.6 & 62.5 & 60.6 & 81.3 & 81.2 & 59.5 & 53.1 & 81.2 & \textbf{62.0} & 58.7 & 65.5 & 73.3 & 51.2 & 58.3 & 60.0 & 69.3 & 61.5 & \textbf{80.0} & 69.7\\
  VGG-16+SCNet-AG+~\cite{SCNet}& 85.5 & 84.4 & 66.3 & 70.8 & 57.4 & 82.7 & 82.3 & 71.6 & \textbf{54.3} & \textbf{95.8} & 55.2 & 59.5 & \textbf{68.6} & 75.0 & 56.3 & 60.4 & 60.0 & \textbf{73.7} & 66.5 & 76.7 & 72.2\\
  VGG-16+CNNGeo~\cite{CNNGeo} & 79.5 & 80.9 & 69.9 & 61.1 & 57.8 & 77.1 & 84.4 & 55.5 & 48.1 & 83.3 & 37.0 & 54.1 & 58.2 & 70.7 & 51.4 & 41.4 & 60.0 & 44.3 & 55.3 & 30.0 & 62.6\\
  ResNet-101+CNNGeo(S)~\cite{CNNGeo} & 83.0 & 82.2 & 81.1 & 50.0 & 57.8 & 79.9 & 92.8 & 77.5 & 44.7 & 85.4 & 28.1 & 69.8 & 65.4 & 77.1 & 64.0 & 65.2 & \textbf{100.0} & 50.8 & 44.3 & 54.4 & 69.5\\
  ResNet-101+CNNGeo(W)~\cite{End-to-end} & 84.7 & 88.9 & 80.9 & 55.6 & 76.6 & 89.5 & \textbf{93.9} & 79.6 & 52.0 & 85.4 & 28.1 & 71.8 & 67.0 & 75.1 & 66.3 & 70.5 & \textbf{100.0} & 62.1 & 62.3 & 61.1 & 74.8\\
  Ours & \textbf{85.6} & \textbf{89.6} & \textbf{82.1} & \textbf{83.3} & \textbf{85.9} & \textbf{92.5} & \textbf{93.9} & \textbf{80.2} & 52.2 & 85.4 & 55.2 & \textbf{75.2} & 64.0 & \textbf{77.9} & 67.2 & \textbf{73.8} & \textbf{100.0} & 65.3 & 69.3 & 61.1 & \textbf{78.0}\\
  \bottomrule
  \end{tabular}
  }
  \end{center}
\end{table}

\subsubsection{Evaluation metric.} 
We evaluate the performance of the proposed method on a semantic correspondence task. To assess the performance, we adopt the percentage of correct keypoints (PCK) metric~\cite{PCK} which measures the percentage of keypoints whose reprojection errors are below the given threshold. The reprojection error is the Euclidean distance $d(\phi(\mathbf{p}), \mathbf{p}^*)$ between the locations of the warped keypoint $\phi(\mathbf{p})$ and the ground truth keypoint $\mathbf{p}^*$. The threshold is defined as $\tau\cdot\max(h,w)$ where $h$ and $w$ are the height and width of the annotated object bounding box on the image, respectively.

\subsubsection{PF-PASCAL~\cite{ProposalFlow}.} \label{data:PF-PASCAL} 
The PF-PASCAL dataset is selected from the PASCAL 2011 keypoint annotations~\cite{PASCAL} containing 1,351 semantically related image pairs from 20 object categories. For images of a category, they contain different object instances of that category with similar poses but different appearances. In addition, the presence of background clutter makes it a challenging dataset on semantic matching. We divide the dataset into 735 pairs for training, 308 pairs for validation, and 308 pairs for testing. Manually annotated correspondences are provided for each image pairs. However, under the weakly supervised setting, we do not use the keypoint annotations for training. The annotations are used only for evaluation. We compute the PCK for each object category with $\tau$ equals to 0.1.

\subsubsection{PF-WILLOW~\cite{ProposalFlow}.} \label{data:PF-WILLOW}
The PF-WILLOW dataset is composed of 100 images with 900 image pairs divided into four semantically related subsets: car, duck, motorbike, and wine bottle. Each subset contains images with large intra-class variations and background clutters. For each image, there are 10 keypoint annotations. We follow Han~\etal~\cite{SCNet} and compute the PCK at three different thresholds with $\tau$ equals to 0.05, 0.1, and 0.15, respectively. 

\subsubsection{TSS~\cite{Taniai}.} \label{data:TSS}
The TSS dataset comprises 400 semantically related image pairs divided into three groups, including FG3DCar, JODS, and PASCAL. FG3DCar contains 195 image pairs of automobiles. JODS is composed of 81 image pairs of airplanes, cars, and horses. There are 124 image pairs of trains, cars, buses, bikes, and motorbikes form the group of PASCAL. Ground truth flows for each image pair are provided. Following Taniai~\etal\cite{Taniai}, we compute the PCK over foreground object by setting $\tau$ to 0.05.

\begin{table}[t]
  \begin{minipage}{0.5\linewidth}
    \scriptsize
    \caption{\textbf{Ablation experiments on PF-PASCAL with $\tau = 0.1$.}}
    \vspace{2pt}
    \label{table:Ablation}
    \centering
    \resizebox{\linewidth}{!}
    {
    \begin{tabular}{l|c}
    \toprule
    Method & mean\\
    \midrule
    Rocco~\etal~\cite{End-to-end} & 74.8\\
    Rocco~\etal~\cite{End-to-end} + $\mathcal{L}_\mathrm{matching}$ & 75.5\\   
    Rocco~\etal~\cite{End-to-end} + $\mathcal{L}_\mathrm{cycle-consis}$ & 77.4\\  
    Rocco~\etal~\cite{End-to-end} + $\mathcal{L}_\mathrm{trans-consis}$ & 77.6\\
    Ours & \textbf{78.0}\\
    \bottomrule
    \end{tabular}
    }
  \end{minipage}
  \begin{minipage}{0.42\linewidth}
    \centering
    \includegraphics[width=\linewidth]{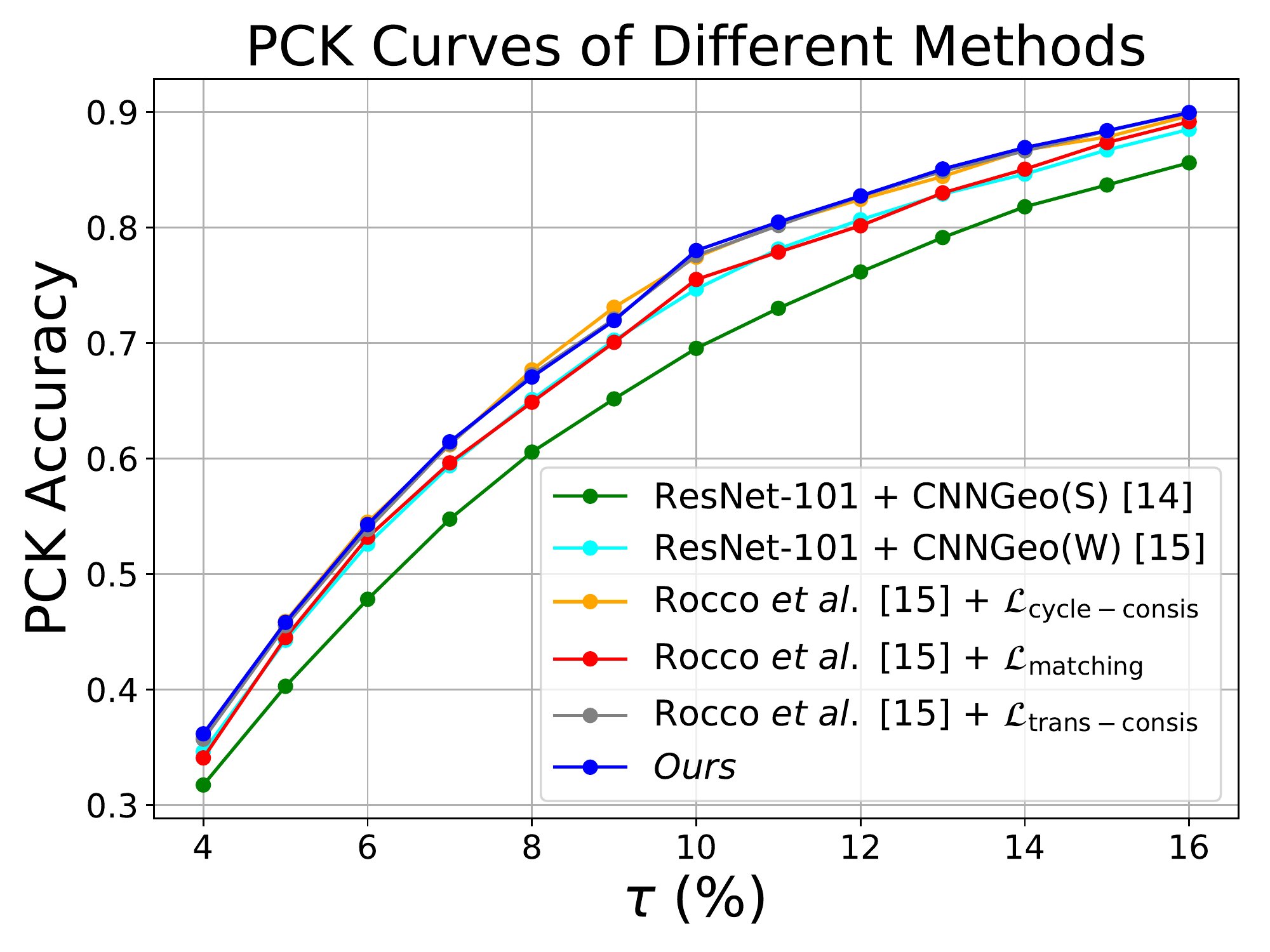}
    \label{fig:PF-PASCAL-PCK}
  \end{minipage}
\end{table}

\setlength{\siximg}{0.16\linewidth}
\begin{figure}[t]
  \centering
  \begin{subfigure}[b]{\siximg}
    \includegraphics[width=1.0\linewidth]{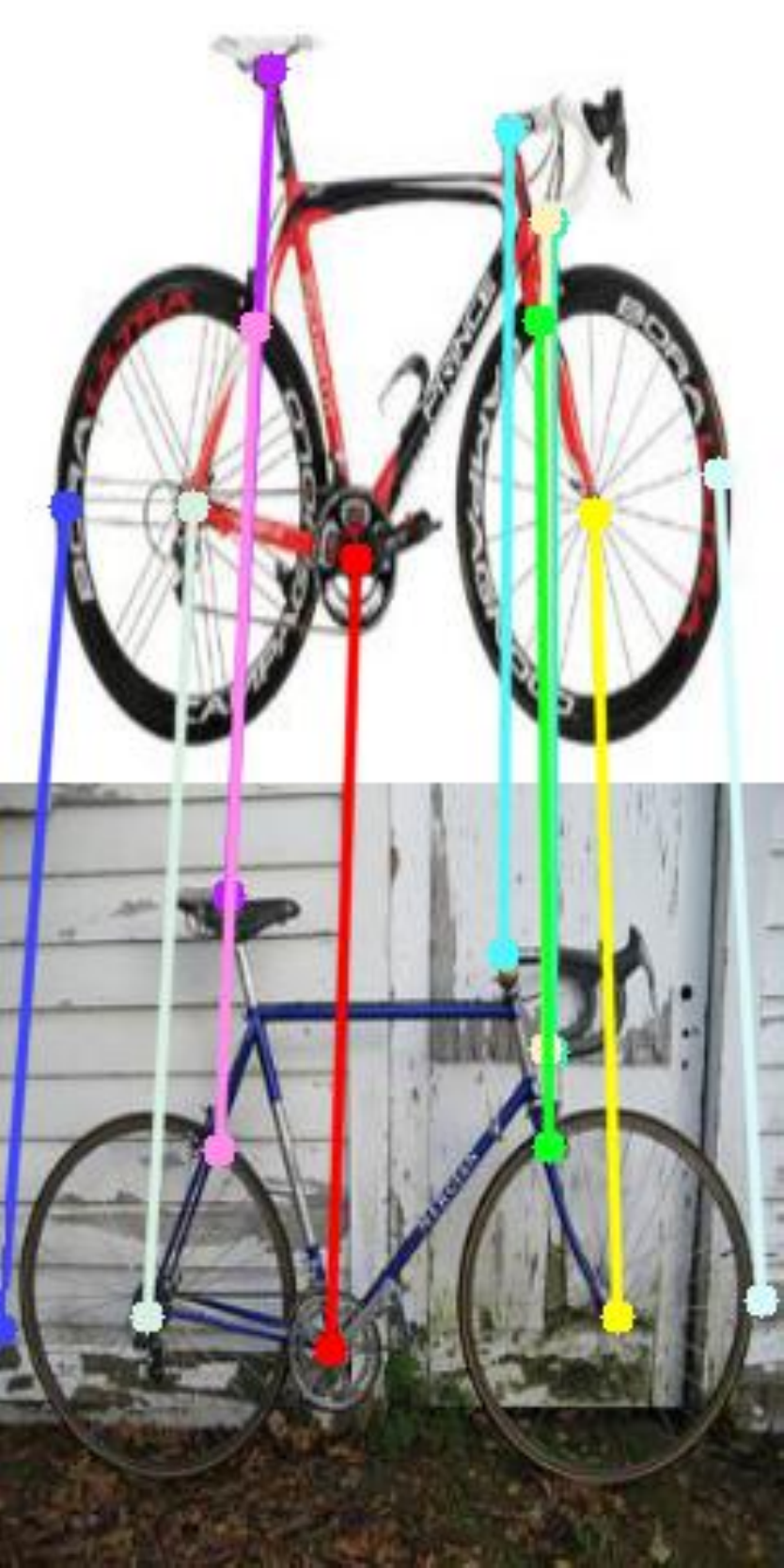}
  \end{subfigure}
  \begin{subfigure}[b]{\siximg}
    \includegraphics[width=1.0\linewidth]{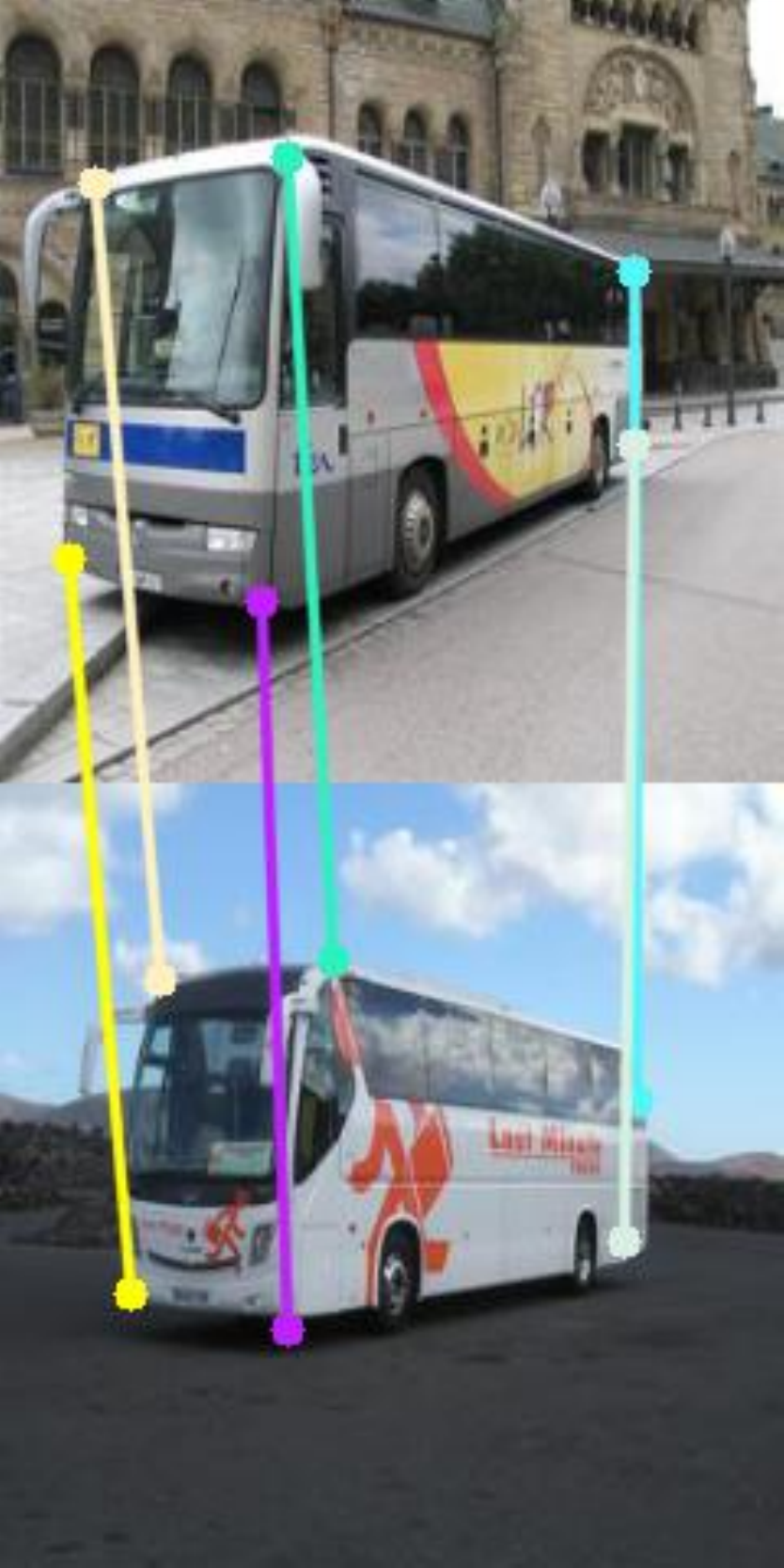}
  \end{subfigure}
  \begin{subfigure}[b]{\siximg}
    \includegraphics[width=1.0\linewidth]{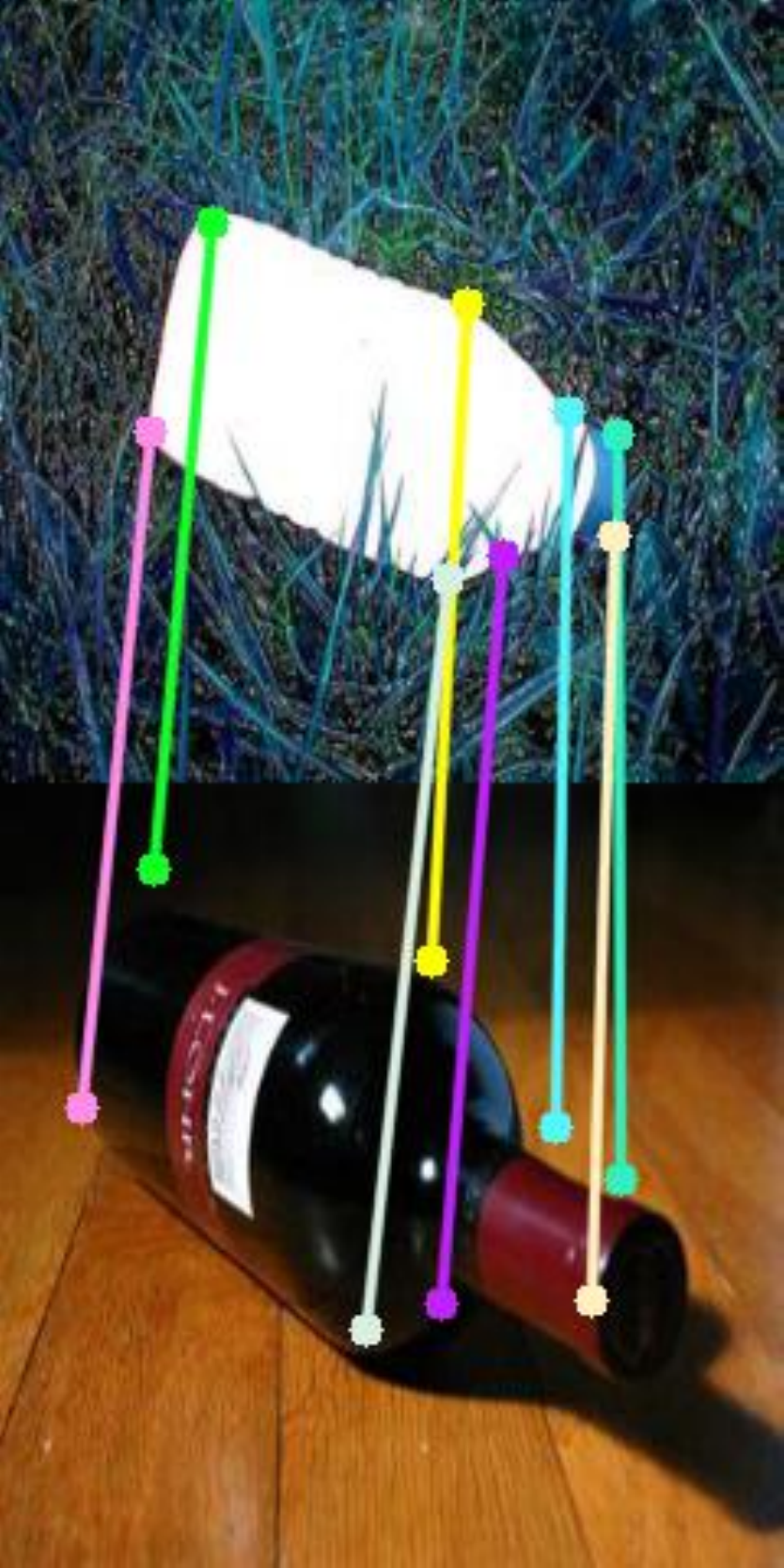}
  \end{subfigure}
  \begin{subfigure}[b]{\siximg}
    \includegraphics[width=1.0\linewidth]{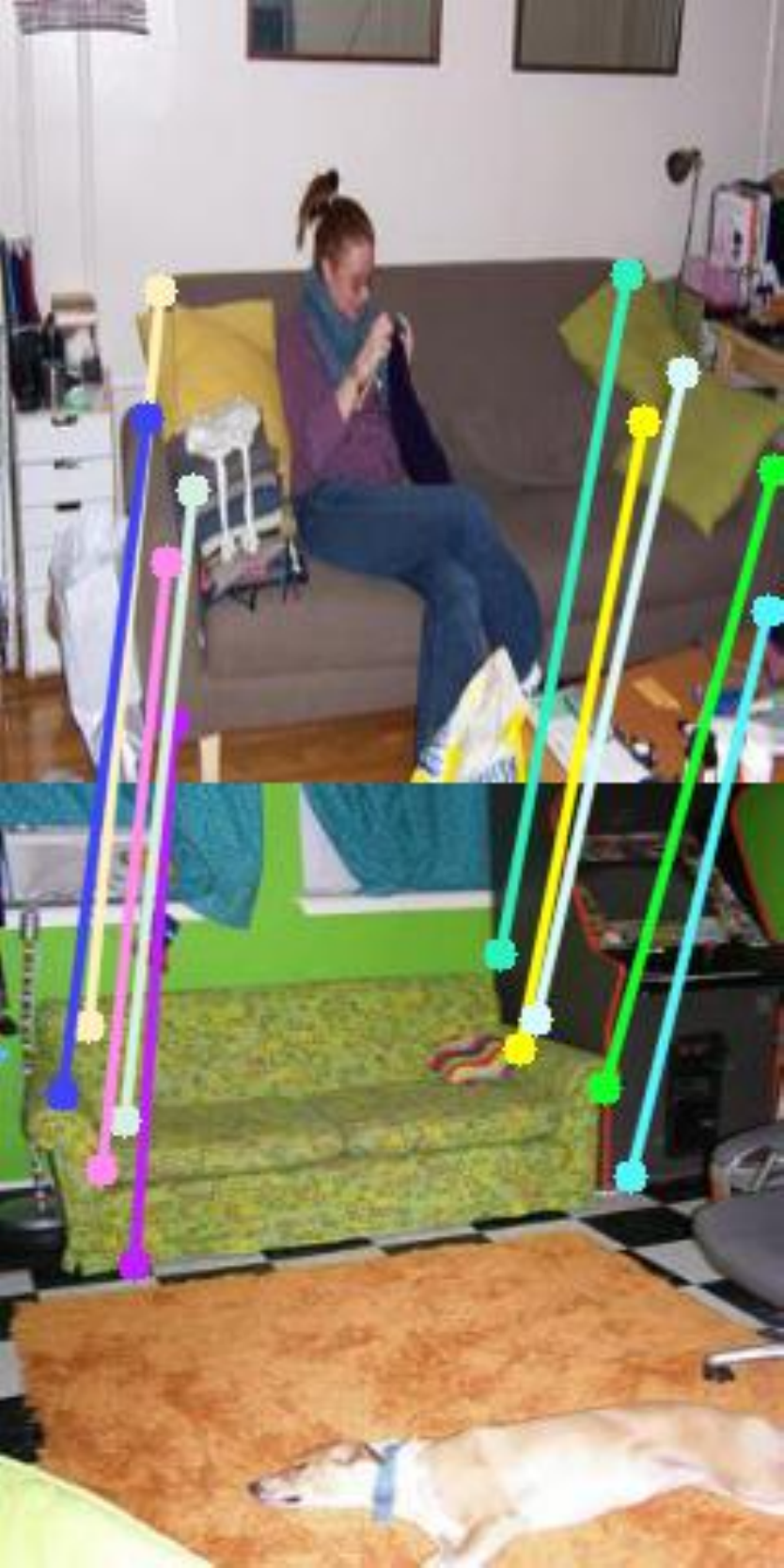}
  \end{subfigure}
  \begin{subfigure}[b]{\siximg}
    \includegraphics[width=1.0\linewidth]{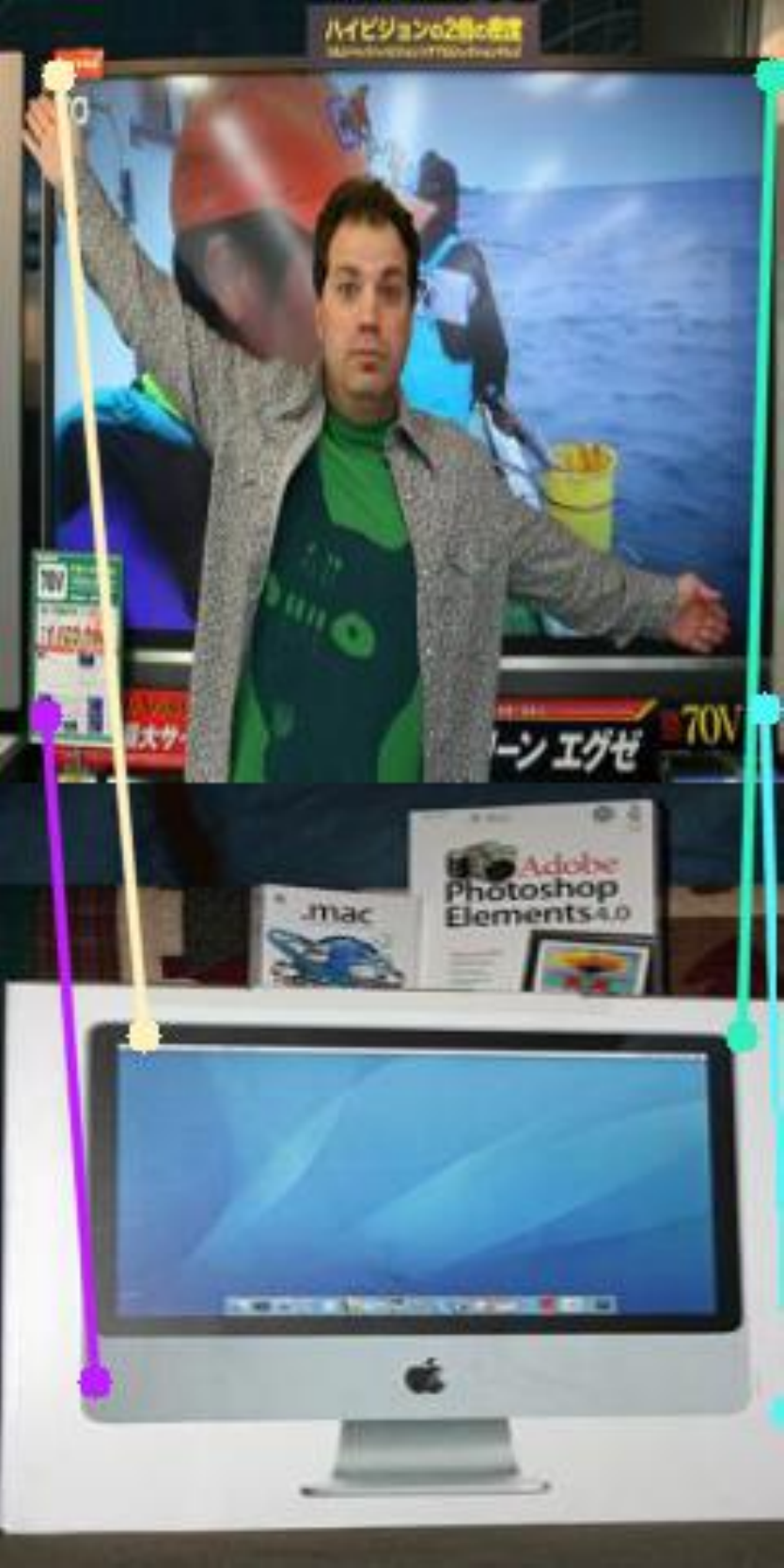}
  \end{subfigure}
  \begin{subfigure}[b]{\siximg}
    \includegraphics[width=1.0\linewidth]{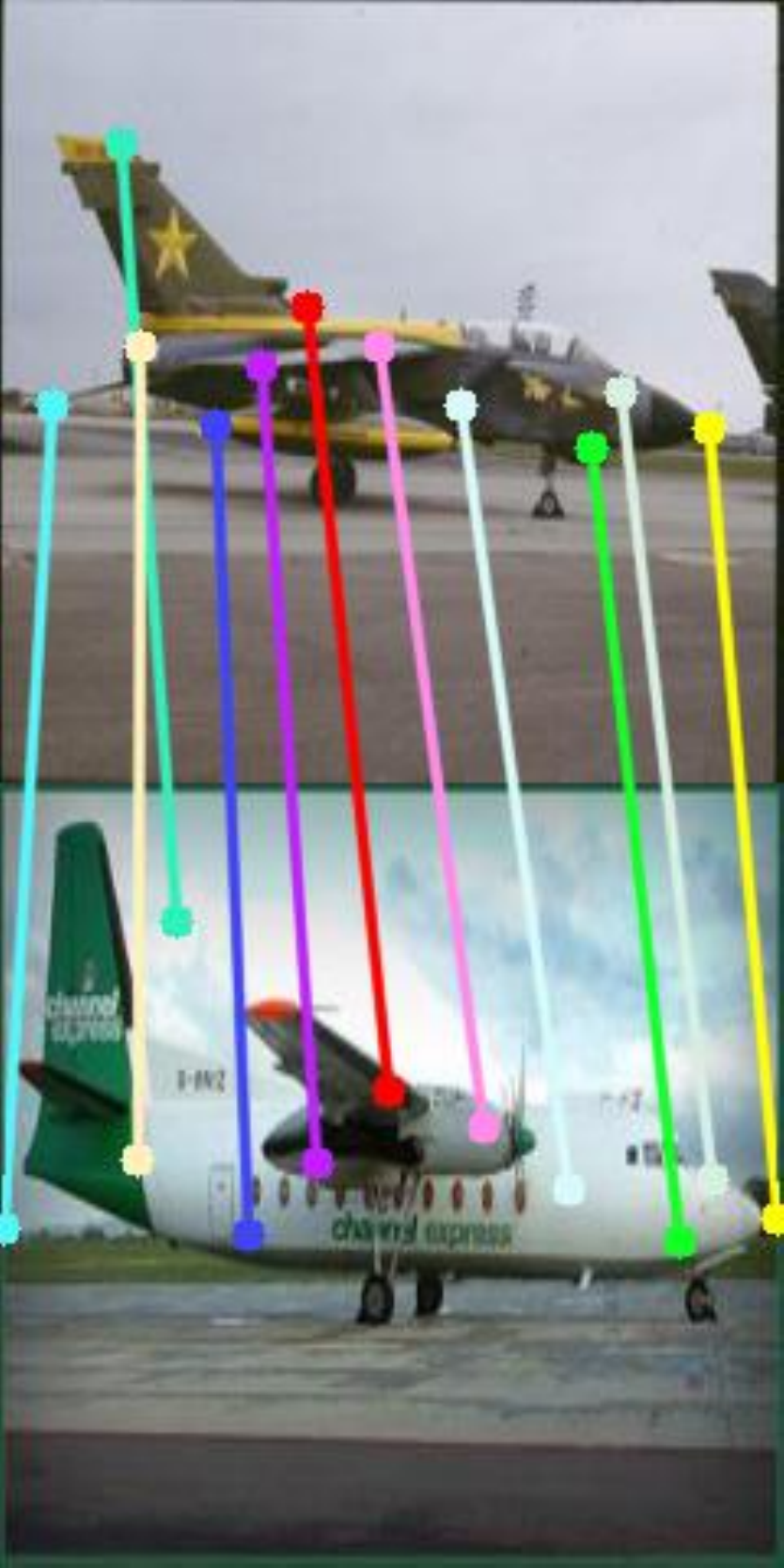}
  \end{subfigure}
  \caption{\textbf{Semantic correspondence results on the PF-PASCAL dataset.} The matched coordinates are linked with color lines.}
  \label{fig:correspondence}
\end{figure}

\subsection{Experimental results on the PF-PASCAL dataset} \label{sec:Results}
In the following, we compare the performance of the proposed method with the state-of-the-art approaches. Note that many of the existing methods require manually annotated correspondences while our model can be trained using only image-level supervision.

\subsubsection{Performance evaluation.}
We compare our method with the Proposal Flow~\cite{ProposalFlow}, the UCN~\cite{UCN}, different versions of the SCNet~\cite{SCNet}, the CNNGeo~\cite{CNNGeo} with different feature extractors, and a weakly supervised approach proposed by Rocco~\etal~\cite{End-to-end}. Table~\ref{table:PF-PASCAL} presents the experimental results for the PF-PASCAL dataset. Our results show that the proposed approach compares favorably against state-of-the-art methods, achieving an overall PCK of 78.0\% (outperforming the previous best method~\cite{End-to-end} by 3.2\%). The advantage of incorporating foreground detection and enforcing cycle consistency constraints can be observed by comparing our method with ResNet-101+CNNGeo(W)~\cite{End-to-end} since both methods utilize the same feature descriptor and are trained with image-level supervision only.
 
Fig.~\ref{fig:correspondence} presents the qualitative results of semantic correspondence on the PF-PASCAL dataset. To further highlight the importance of each component of the proposed method, we present an ablation study of our method.

\setlength{\fiveimg}{0.19\linewidth}
\begin{figure}[t]
  \centering
  \begin{subfigure}[t]{\fiveimg}
    \includegraphics[width=1.0\linewidth]{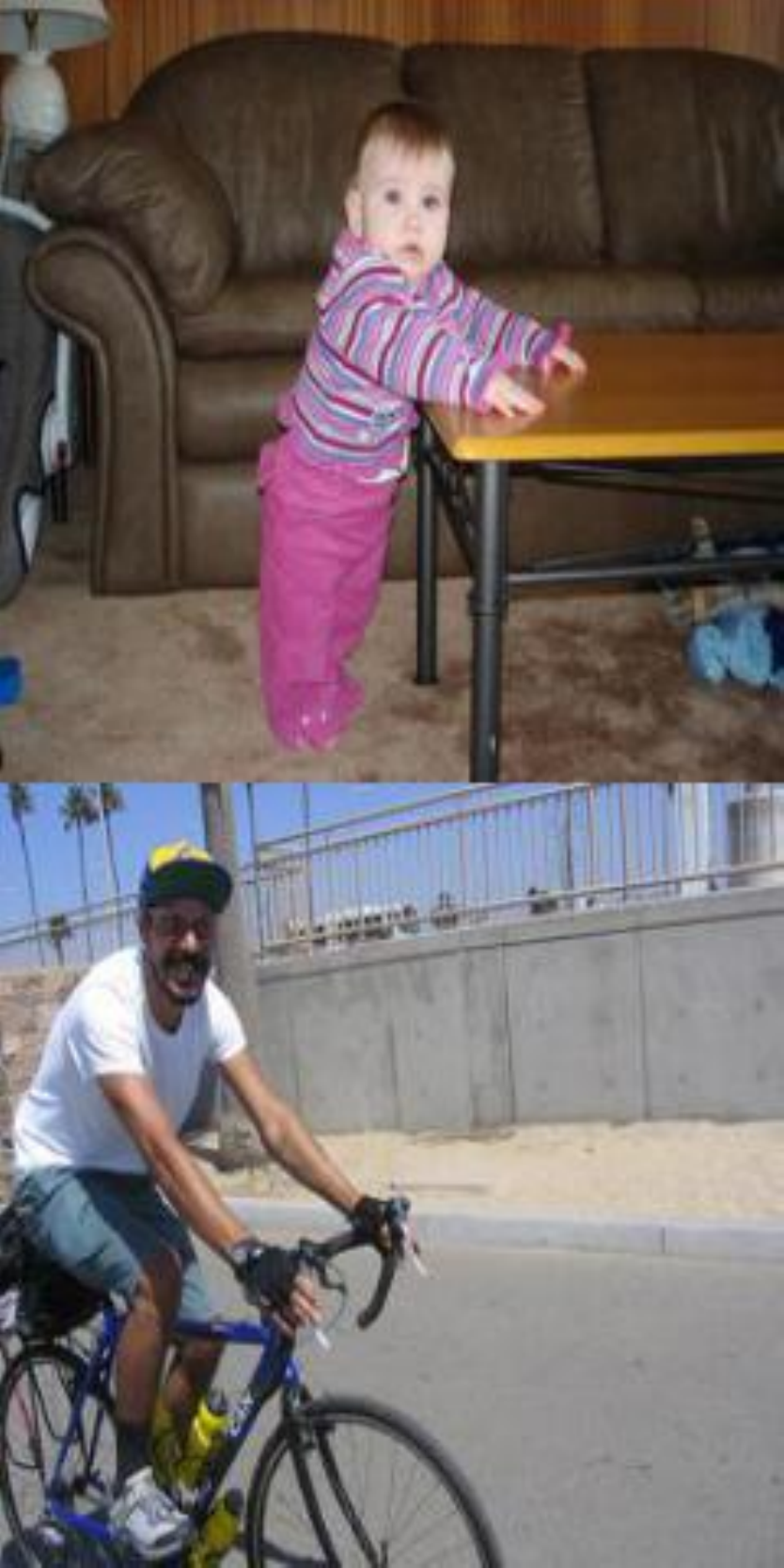}
    \caption*{An image pair}
    \label{fig:fig1-a}
  \end{subfigure}
  \begin{subfigure}[t]{\fiveimg}
    \includegraphics[width=1.0\linewidth]{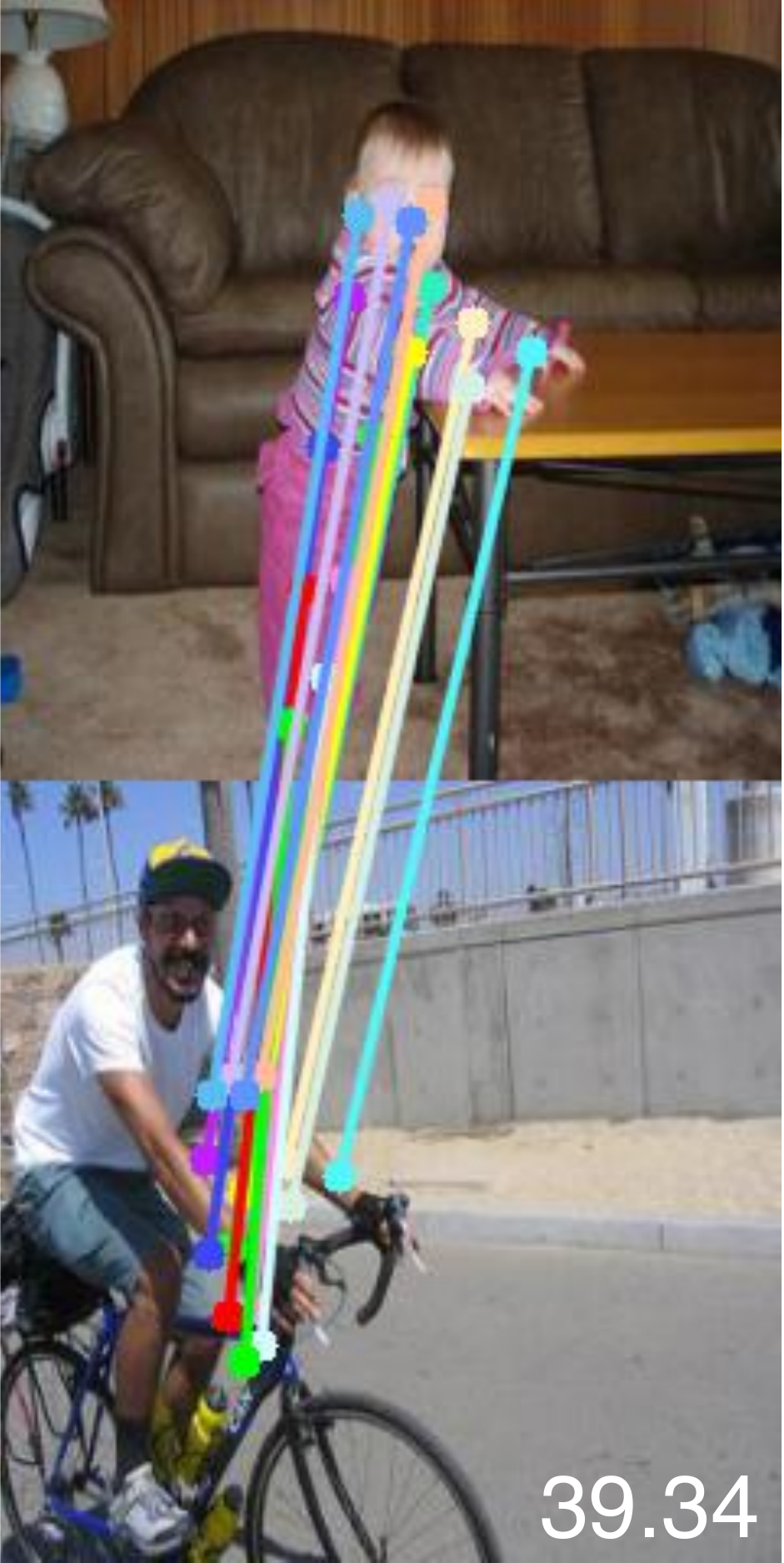}
    \caption*{Rocco~\etal~\cite{End-to-end}}
    \label{fig:fig1-b}
  \end{subfigure}
  \begin{subfigure}[t]{\fiveimg}
    \includegraphics[width=1.0\linewidth]{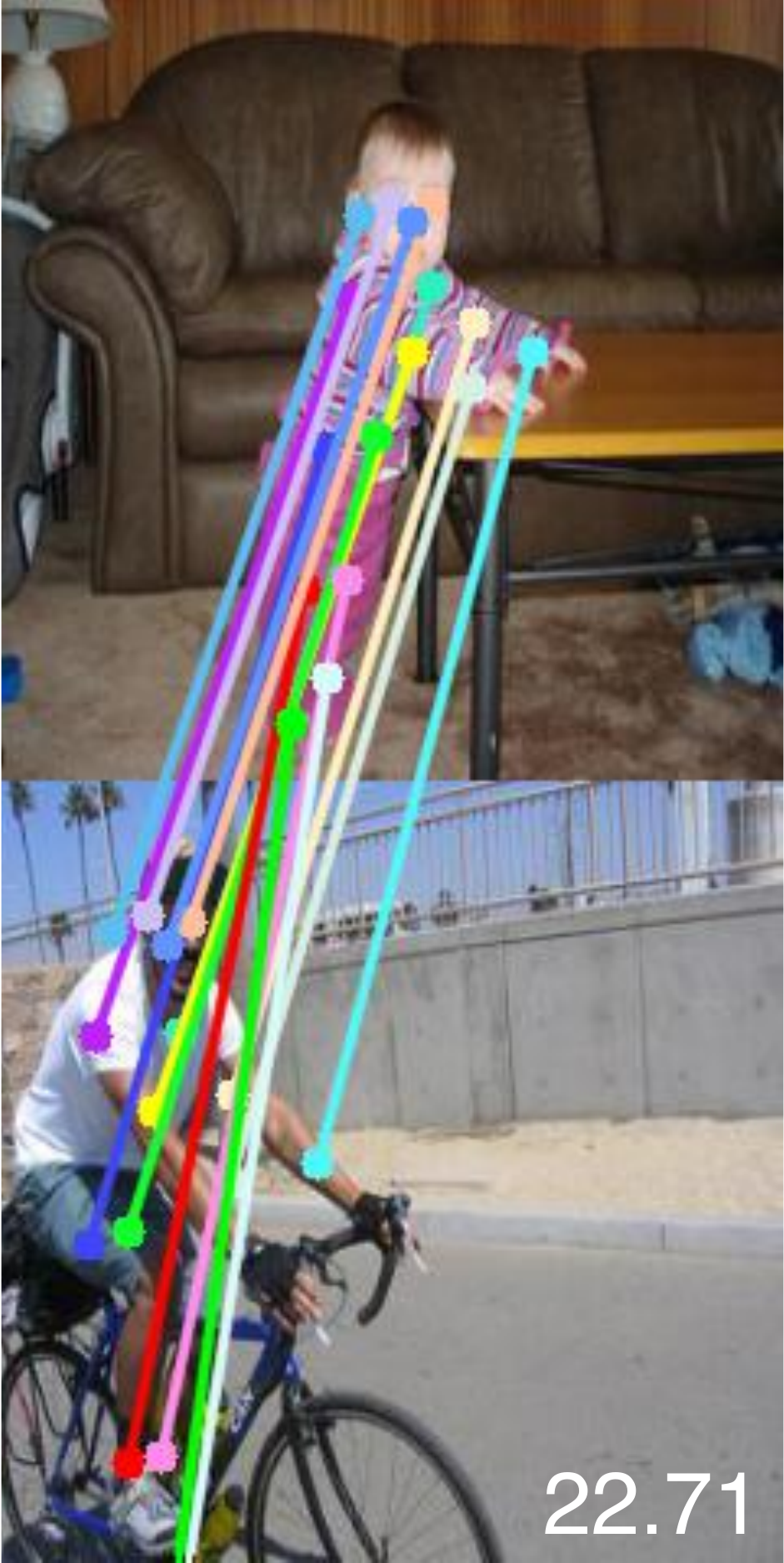}
    \caption*{Rocco~\etal~\cite{End-to-end} \\ + foreground \\ detection}
    \label{fig:fig1-c}
  \end{subfigure}
  \begin{subfigure}[t]{\fiveimg}
    \includegraphics[width=1.0\linewidth]{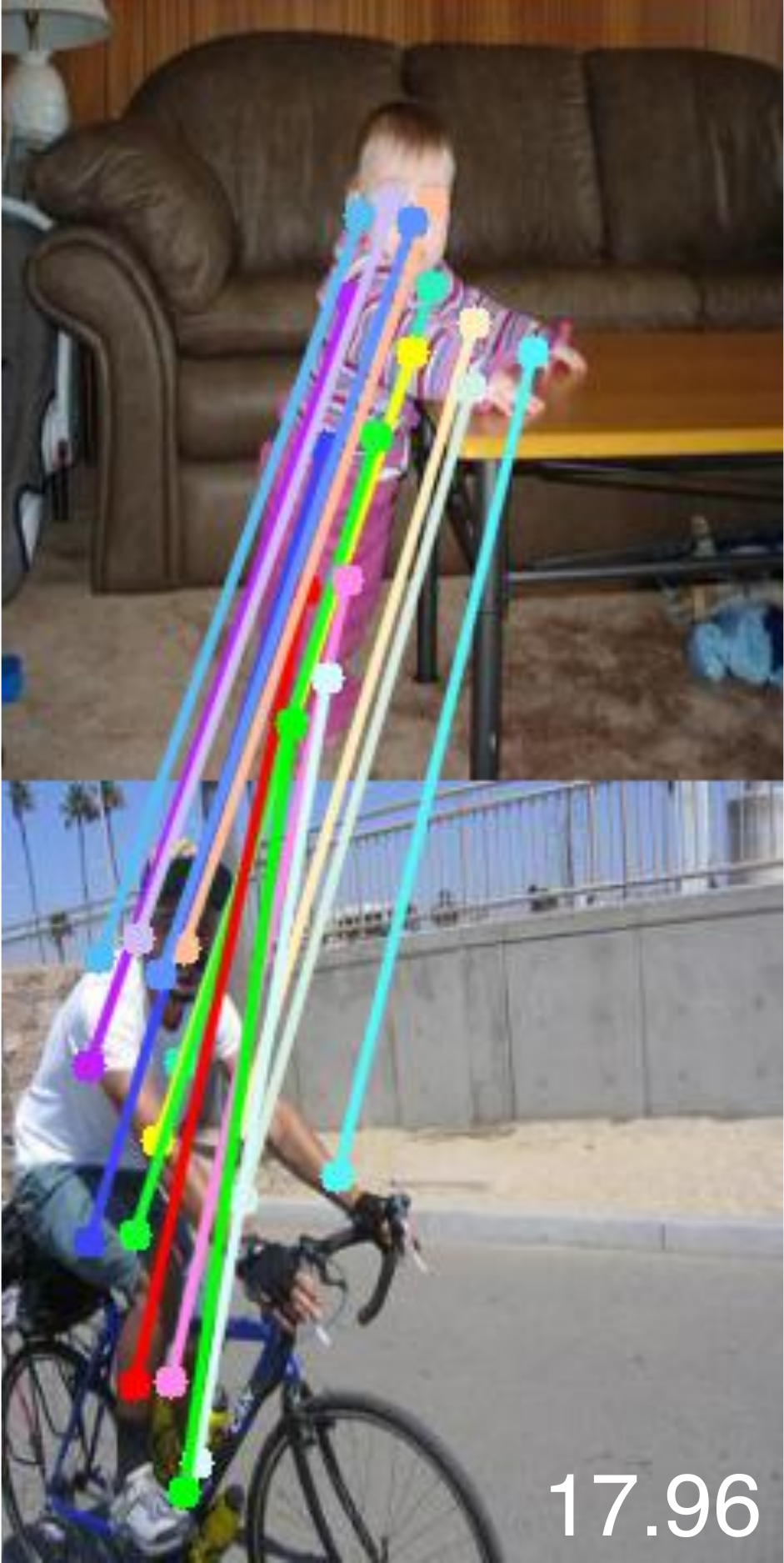}
    \caption*{Rocco~\etal~\cite{End-to-end} \\ + cycle \\ consistency}
    \label{fig:fig1-d}
  \end{subfigure}
  \begin{subfigure}[t]{\fiveimg}
    \includegraphics[width=1.0\linewidth]{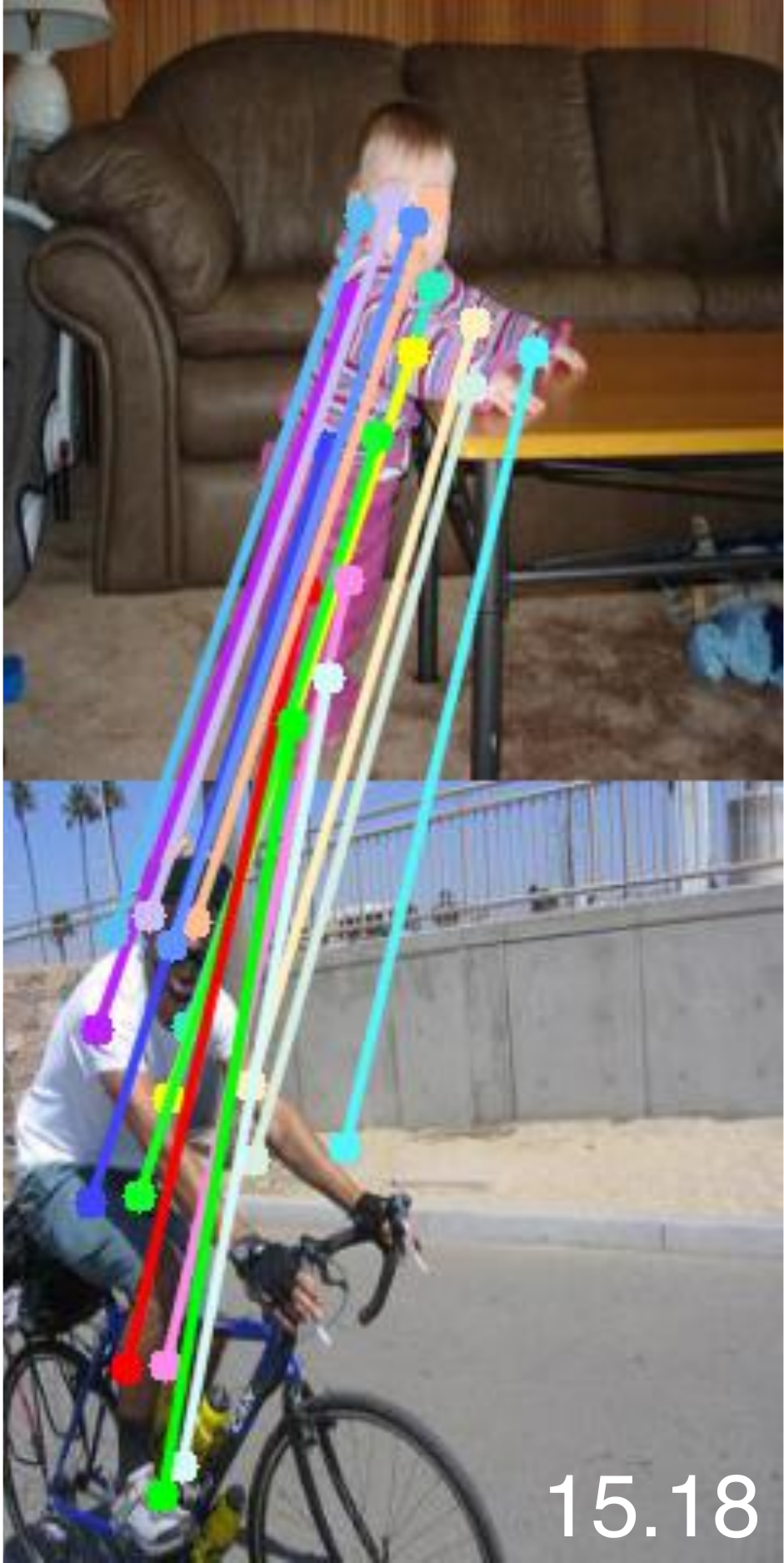}
    \caption*{Ours}
    \label{fig:fig1-e}
  \end{subfigure}
  \caption{\textbf{Visualization of the effect of each component.} Given an image pair, existing methods often suffer from the negative impacts due to background clutter. Integrating foreground detection into semantic matching alleviates the unfavorable effects of background clutters. Enforcing cycle consistency improves the matching result. Our method integrates foreground detection and exploits cycle-consistency property in semantic matching, resulting in more accurate results. The bottom right corners display the errors, namely the average distances between the predicted keypoints and the ground truth.}
  \label{fig:vis-abla}
\end{figure}

\setlength{\threeimg}{0.3\linewidth}
\begin{figure}[t]
  \centering
  \begin{subfigure}[b]{\threeimg}
    \centering\includegraphics[width=1.0\textwidth]{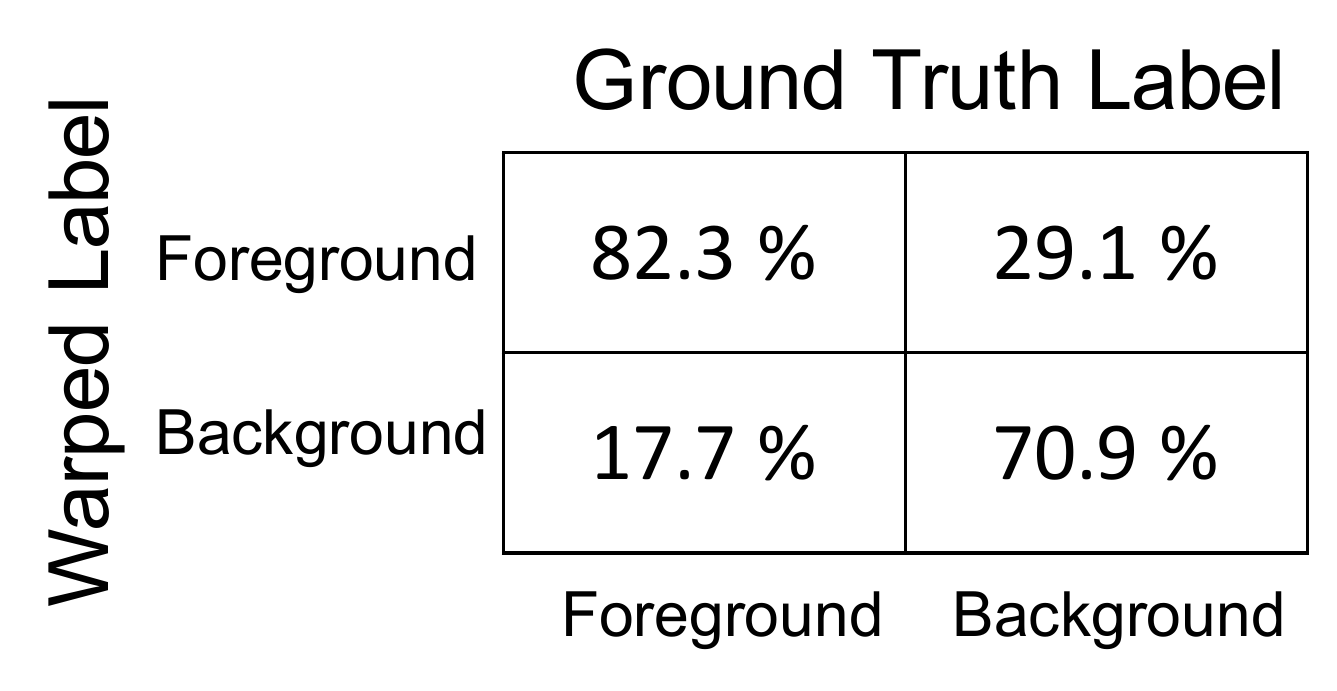}
    \vspace{\figmargin}
    \caption*{Rocco~\etal~\cite{End-to-end}}
  \end{subfigure}
  \begin{subfigure}[b]{\threeimg}
    \centering\includegraphics[width=1.0\textwidth]{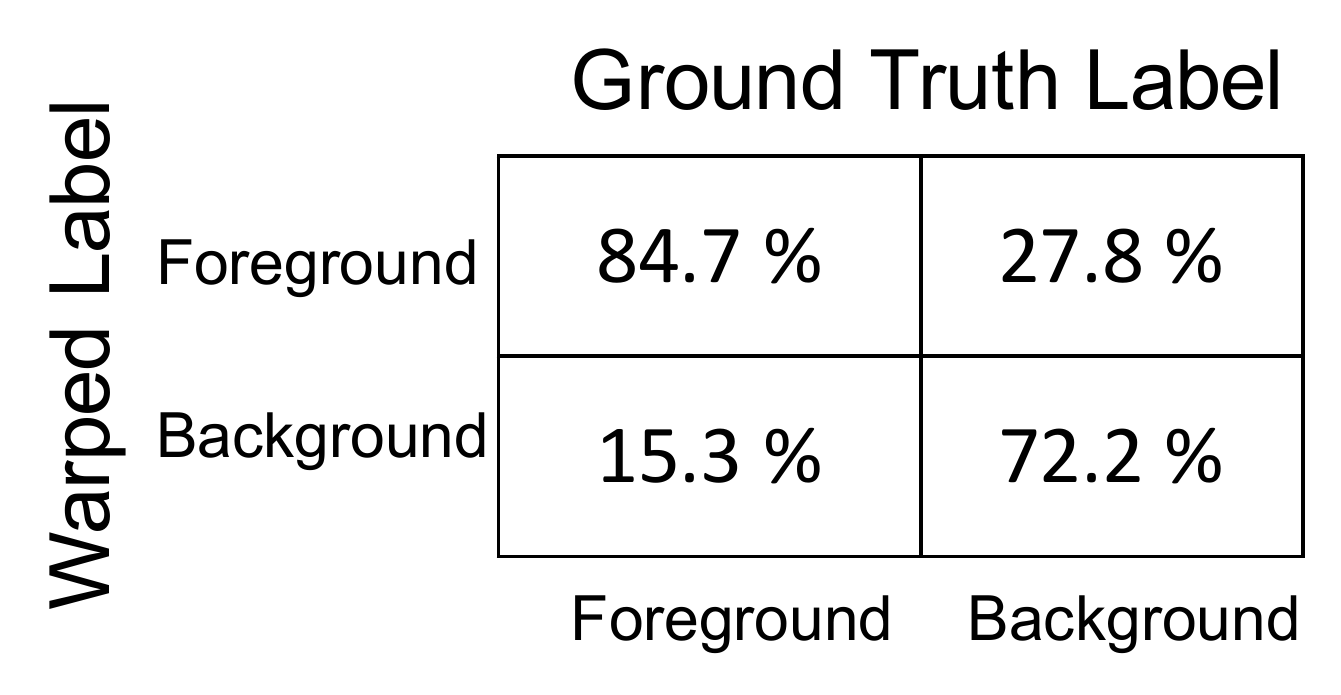}
    \vspace{\figmargin}
    \caption*{Ours}
  \end{subfigure}
  \begin{subfigure}[b]{\threeimg}
    \centering\includegraphics[width=1.0\textwidth]{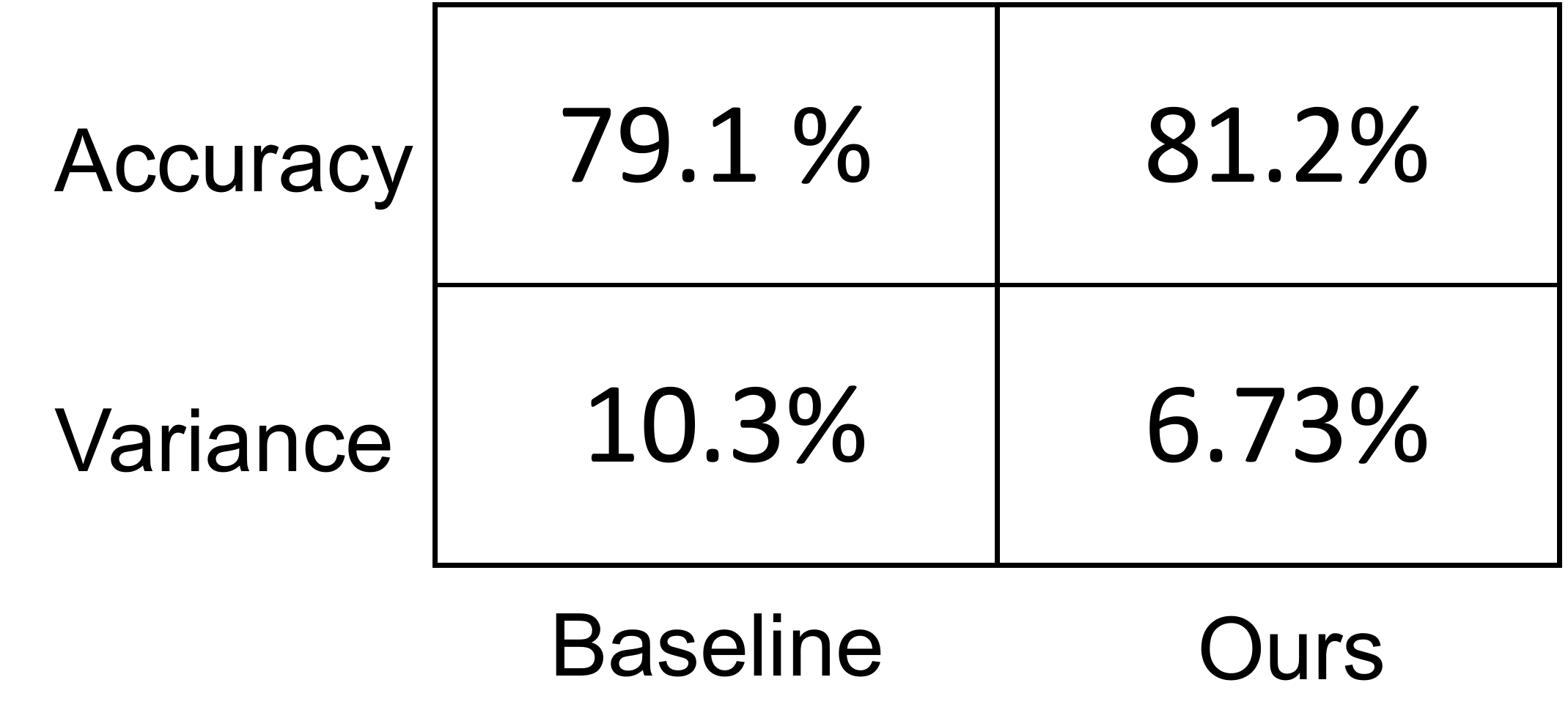}
    \vspace{\figmargin}
    \caption*{Comparison}
  \end{subfigure}
  \caption{Effect of using the foreground-guided matching loss $\mathcal{L}_\mathrm{matching}$.}
  \label{LM}
\end{figure}

\begin{table}[t]
  \begin{minipage}{.45\linewidth}
  \scriptsize
  \caption{\small\textbf{Results on PF-WILLOW.}}
  \label{table:PF-WILLOW}
  \centering
  \resizebox{\textwidth}{!} 
  {
  \begin{tabular}{l | c | c | c}
  \toprule
  Method & $\tau$ = 0.05 & $\tau$ = 0.1 & $\tau$ = 0.15 \\ [1ex]
  \midrule
  SIFT Flow~\cite{SIFTFlow} & 0.247 & 0.380 & 0.504\\
  DAISY w/SF~\cite{DAISY} & 0.324 & 0.456 & 0.555\\
  DeepC w/SF~\cite{DeepC} & 0.212 & 0.364 & 0.518\\
  LIFT w/SF~\cite{Lift} & 0.224 & 0.346 & 0.489\\
  VGG w/SF~\cite{VGG} & 0.324 & 0.456 & 0.555\\
  FCSS w/SF~\cite{FCSS} & 0.354 & 0.532 & 0.681\\
  LOM HOG~\cite{ProposalFlow} & 0.284 & 0.568 & 0.682\\
  UCN~\cite{UCN} & 0.291 & 0.417 & 0.513\\
  DSFM~\cite{ufer2017deep} & - & 0.680 & - \\
  SCNet-A~\cite{SCNet} & 0.390 & 0.725 & 0.873\\
  SCNet-AG~\cite{SCNet} & 0.394 & 0.721 & 0.871\\
  SCNet-AG+~\cite{SCNet} & 0.386 & 0.704 & 0.853\\
  ResNet-101+CNNGeo(S)~\cite{CNNGeo} & 0.448 & 0.777 & 0.899\\
  ResNet-101+CNNGeo(W)~\cite{End-to-end} & 0.477 & 0.812 & 0.917\\
  Ours & \textbf{0.491} & \textbf{0.819} & \textbf{0.922}\\
  \bottomrule
  \end{tabular}
  }
  \end{minipage}
  \begin{minipage}{.53\linewidth}
  \scriptsize
  \caption{\textbf{Results on TSS.} Marker $^*$ indicates that the method uses extra images from the PASCAL VOC 2007 dataset.}
  \label{table:TSS}
  \centering
  \resizebox{\textwidth}{!} 
  {
  \begin{tabular}{l | c | c | c | c}
  \toprule
  Method & FG3DCar & JODS & PASCAL & Avg.\\ [1ex]
  \midrule
  HOG+PF-LOM~\cite{ProposalFlow} & 0.786 & 0.653 & 0.531 & 0.657\\
  HOG+TSS~\cite{Taniai} & 0.830 & 0.595 & 0.483 & 0.636\\
  FCSS+SIFT Flow~\cite{FCSS} & 0.830 & 0.656 & 0.494 & 0.660\\
  FCSS+PF-LOM~\cite{FCSS} & 0.839 & 0.635 & 0.582 & 0.685\\
  HOG+OADSC~\cite{ObjAware}$^*$ & 0.875 & 0.708 & 0.729 & 0.771\\
  FCSS+DCTM~\cite{DCTM} & 0.891 & 0.721 & \textbf{0.610} & 0.740\\
  VGG-16+CNNGeo~\cite{CNNGeo} & 0.835 & 0.656 & 0.527 & 0.673\\
  ResNet-101+CNNGeo(S)~\cite{CNNGeo} & 0.886 & 0.758 & 0.560 & 0.735\\
  ResNet-101+CNNGeo(W)~\cite{End-to-end} & 0.892 & 0.758 & 0.562 & 0.737\\
  Ours & \textbf{0.898} & \textbf{0.768} & 0.560 & \textbf{0.742}\\
  \bottomrule
  \end{tabular}
  }
  \end{minipage} 
\end{table}

\setlength{\fourimg}{0.24\linewidth}
\begin{figure}[t]
  \centering
  \begin{subfigure}[b]{\fourimg}
    \includegraphics[width=1.0\linewidth]{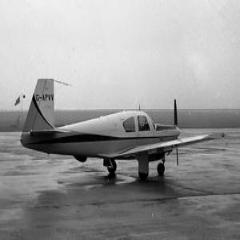}
    \caption*{Source image}
  \end{subfigure}
  \begin{subfigure}[b]{\fourimg}
    \includegraphics[width=1.0\linewidth]{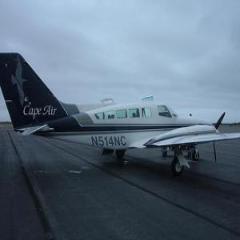}
    \caption*{Target image}
  \end{subfigure}
  \begin{subfigure}[b]{\fourimg}
    \includegraphics[width=1.0\linewidth]{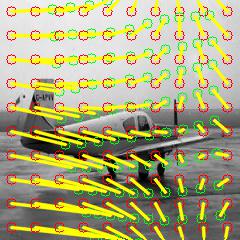}
    \caption*{Rocco~\etal~\cite{End-to-end}}
  \end{subfigure}
  \begin{subfigure}[b]{\fourimg}
    \includegraphics[width=1.0\linewidth]{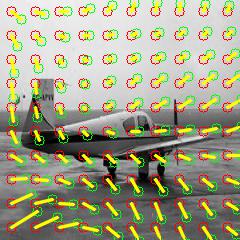}
    \caption*{Ours}
  \end{subfigure}
  \caption{\textbf{Cycle consistency property.} We present the visualization that demonstrates the effect of forward-backward consistency loss where the red points indicate the keypoints while the green points denote the reprojected points. Yellow line represents the distance (loss) between the linked points.}
  \label{fig:F-B}
\end{figure} 

\subsubsection{Ablation study.} \label{sec:AblationStudy}
To analyze the importance of each loss function, we conduct ablation experiments on the PF-PASCAL~\cite{ProposalFlow} dataset. Table~\ref{table:Ablation} presents the mean PCK value of the variants of our approach evaluated on the PF-PASCAL dataset with $\tau$ equals to $0.1$. Our results show that both $\mathcal{L}_\mathrm{cycle-consis}$ and $\mathcal{L}_\mathrm{trans-consis}$ substantially improve the performance when comparing with Rocco~\etal~\cite{End-to-end}. We visualize the effect of each component in Fig.~\ref{fig:vis-abla}. To demonstrate the effectiveness of forward-backward consistency property, we visualize an example in Fig.~\ref{fig:F-B} where the red points indicate the key points and the green points represent the reprojected points. The length of the yellow line represents the distance (loss) between the corresponding points. We observe that enforcing cycle consistency effectively encourages the network to produce geometrically consistent predictions. However, the performance gain of using only the foreground-guided matching loss $\mathcal{L}_\mathrm{matching}$ is modest. We believe that the reason is due to the evaluation protocol of datasets considers only the matching on the foreground region. Namely, matching a background pixel in the source image to a foreground pixel in the target image will not be penalized. To demonstrate the effect of foreground-guided matching loss $\mathcal{L}_\mathrm{matching}$, we compute the percentage of correctly warped pixels (\ie pixels in the foreground/background regions that are correctly warped into foreground/background region) over the entire dataset. As shown in Fig.~\ref{LM}, our method effectively reduces the errors in matching pixels from foreground to background and vice versa. The improvement here is important in real-world applications but is not reflected in the metric used in the standard datasets. We also note that our method may not produce a clear figure-ground separation when the background contains visually similar regions to the foreground object in the other image. However, this case is also challenging for most methods.

The ablation study shows that all of the proposed components play crucial roles in producing accurate matching results. From Fig.~\ref{fig:PF-PASCAL-PCK}, we observe that the proposed method outperforms the best competitor~\cite{End-to-end} with a significant margin at multiple thresholds.

\subsection{Experimental results on the PF-WILLOW and TSS datasets}
To evaluate the generalization capability, we apply the learned model trained on the PF-PASCAL dataset to test directly on the PF-WILLOW and TSS datasets without finetuning on these two datasets.

\subsubsection{Results on the PF-WILLOW dataset.}
Table~\ref{table:PF-WILLOW} presents the quantitative results for the PF-WILLOW dataset. We compare the performance with several recent methods~\cite{UCN,FCSS,SCNet,CNNGeo,End-to-end} as well as conventional approaches~\cite{SIFTFlow,DAISY,DeepC,Lift,VGG} using hand-crafted features. The results are directly taken from~\cite{SCNet} except \cite{CNNGeo,End-to-end}. For \cite{CNNGeo} and \cite{End-to-end}, we run the code provided by the authors to obtain the results. From Table \ref{table:PF-WILLOW}, we observe that our model achieves the state-of-the-art performance with all three thresholds.

\subsubsection{Results on the TSS dataset.}
We also evaluate the performance on the TSS dataset. Table~\ref{table:TSS} presents the quantitative results. We observe that our method achieves the state-of-the-art performance on two of the three groups of the TSS dataset: FG3DCar and JODS. Our results are slightly worse than that in \cite{ObjAware} in the PASCAL group. However, the method in \cite{ObjAware} uses additional images from the PASCAL VOC 2007 dataset. We report their results for completeness. Under the same experimental settings, the proposed method performs favorably against existing approaches.

\section{Conclusions}  \label{sec:Conclusion}
In this work, we present an effective approach to improve semantic matching. The core technical novelty of our approach lies in the explicit modeling of a \emph{foreground detection} module to suppress the effect of background clutter and exploiting the \emph{cycle consistency} constraints so that the predicted geometric transformations are geometrically plausible and consistent across multiple images. The network training requires only training image pairs with image-level supervision and thus significantly alleviates the cost of constructing and labeling large-scale training datasets. Experimental results demonstrate that our approach performs favorably against existing semantic matching algorithms on several standard benchmarks. Moving forward, we believe that the semantic matching network can be further integrated to other computer vision tasks, e.g., supporting 3D semantic object reconstruction and fine-grained visual recognition.

{\flushleft {\bf Acknowledgement.}}
This work is supported in part by Ministry of Science and Technology under grants MOST 105-2221-E-001-030-MY2 and MOST 107-2628-E-001-005-MY3.

\bibliographystyle{splncs04}
\bibliography{reference}

\begin{thebibliography}{10}
\providecommand{\url}[1]{\texttt{#1}}
\providecommand{\urlprefix}{URL }
\providecommand{\doi}[1]{https://doi.org/#1}

\bibitem{PASCAL}
Bourdev, L., Malik, J.: Poselets: Body part detectors trained using 3d human
  pose annotations. In: ICCV (2009)

\bibitem{chen2015co}
Chen, H.Y., Lin, Y.Y., Chen, B.Y.: Co-segmentation guided hough transform for
  robust feature matching. TPAMI  (2015)

\bibitem{chen2019learning}
Chen, Y.C., Li, Y.J., Du, X., Wang, Y.C.F.: Learning resolution-invariant deep
  representations for person re-identification. In: AAAI (2019)

\bibitem{chen2019crdoco}
Chen, Y.C., Lin, Y.Y., Yang, M.H., Huang, J.B.: Crdoco: Pixel-level domain
  transfer with cross-domain consistency. In: CVPR (2019)

\bibitem{chen2019show}
Chen, Y.C., Lin, Y.Y., Yang, M.H., Huang, J.B.: Show, match and segment: Joint
  weakly supervised learning of semantic matching and object co-segmentation.
  TPAMI  (2020)

\bibitem{UCN}
Choy, C.B., Gwak, J., Savarese, S., Chandraker, M.: Universal correspondence
  network. In: NIPS (2016)

\bibitem{HoG}
Dalal, N., Triggs, B.: Histograms of oriented gradients for human detection.
  In: CVPR (2005)

\bibitem{ProposalFlow}
Ham, B., Cho, M., Schmid, C., Ponce, J.: Proposal flow: Semantic
  correspondences from object proposals. TPAMI  (2017)

\bibitem{SCNet}
Han, K., Rezende, R.S., Ham, B., Wong, K.Y.K., Cho, M., Schmid, C., Ponce, J.:
  Scnet: Learning semantic correspondence. In: ICCV (2017)

\bibitem{ResNet}
He, K., Zhang, X., Ren, S., Sun, J.: Deep residual learning for image
  recognition. In: CVPR (2016)

\bibitem{horn1981determining}
Horn, B.K., Schunck, B.G.: Determining optical flow. Artificial intelligence
  (1981)

\bibitem{hsuco}
Hsu, K.J., Lin, Y.Y., Chuang, Y.Y.: Co-attention cnns for unsupervised object
  co-segmentation. In: IJCAI (2018)

\bibitem{hsu2015robust}
Hsu, K.J., Lin, Y.Y., Chuang, Y.Y., et~al.: Robust image alignment with
  multiple feature descriptors and matching-guided neighborhoods. In: CVPR
  (2015)

\bibitem{hu2016progressive}
Hu, Y.T., Lin, Y.Y.: Progressive feature matching with alternate descriptor
  selection and correspondence enrichment. In: CVPR (2016)

\bibitem{hu2015matching}
Hu, Y.T., Lin, Y.Y., Chen, H.Y., Hsu, K.J., Chen, B.Y.: Matching images with
  multiple descriptors: An unsupervised approach for locally adaptive
  descriptor selection. TIP  (2015)

\bibitem{huang2016temporally}
Huang, J.B., Kang, S.B., Ahuja, N., Kopf, J.: Temporally coherent completion of
  dynamic video. ACM Transactions on Graphics (TOG)  (2016)

\bibitem{WarpNet}
Kanazawa, A., Jacobs, D.W., Chandraker, M.: Warpnet: Weakly supervised matching
  for single-view reconstruction. In: CVPR (2016)

\bibitem{Deformable}
Kim, J., Liu, C., Sha, F., Grauman, K.: Deformable spatial pyramid matching for
  fast dense correspondences. In: CVPR (2013)

\bibitem{FCSS}
Kim, S., Min, D., Ham, B., Jeon, S., Lin, S., Sohn, K.: Fcss: Fully
  convolutional self-similarity for dense semantic correspondence. In: CVPR
  (2017)

\bibitem{DCTM}
Kim, S., Min, D., Lin, S., Sohn, K.: Dctm: Discrete-continuous transformation
  matching for semantic flow. In: CVPR (2017)

\bibitem{Adam}
Kingma, D.P., Ba, J.: Adam: A method for stochastic optimization. In: ICLR
  (2015)

\bibitem{lai2018learning}
Lai, W.S., Huang, J.B., Wang, O., Shechtman, E., Yumer, E., Yang, M.H.:
  Learning blind video temporal consistency. In: ECCV (2018)

\bibitem{DRIT}
Lee, H.Y., Tseng, H.Y., Huang, J.B., Singh, M., Yang, M.H.: Diverse
  image-to-image translation via disentangled representations. In: ECCV (2018)

\bibitem{li2019recover}
Li, Y.J., Chen, Y.C., Lin, Y.Y., Du, X., Wang, Y.C.F.: Recover and identify: A
  generative dual model for cross-resolution person re-identification. In: ICCV
  (2019)

\bibitem{li2020cross}
Li, Y.J., Chen, Y.C., Lin, Y.Y., Wang, Y.C.F.: Cross-resolution adversarial
  dual network for person re-identification and beyond. arXiv  (2020)

\bibitem{SIFTFlow}
Liu, C., Yuen, J., Torralba, A.: Sift flow: Dense correspondence across scenes
  and its applications. TPAMI  (2011)

\bibitem{lucas1981iterative}
Lucas, B.D., Kanade, T., et~al.: An iterative image registration technique with
  an application to stereo vision. In: IJCAI (1981)

\bibitem{UnFlow}
Meister, S., Hur, J., Roth, S.: Unflow: Unsupervised learning of optical flow
  with a bidirectional census loss. In: AAAI (2018)

\bibitem{mustafa}
Mustafa, A., Hilton, A.: Semantically coherent co-segmentation and
  reconstruction of dynamic scenes. In: CVPR (2017)

\bibitem{AnchorNet}
Novotny, D., Larlus, D., Vedaldi, A.: Anchornet: A weakly supervised network to
  learn geometry-sensitive features for semantic matching. In: CVPR (2017)

\bibitem{CNNGeo}
Rocco, I., Arandjelovi{\'c}, R., Sivic, J.: Convolutional neural network
  architecture for geometric matching. In: CVPR (2017)

\bibitem{End-to-end}
Rocco, I., Arandjelovi{\'c}, R., Sivic, J.: End-to-end weakly-supervised
  semantic alignment. In: CVPR (2018)

\bibitem{StereoMatching}
Scharstein, D., Szeliski, R.: A taxonomy and evaluation of dense two-frame
  stereo correspondence algorithms. IJCV  (2002)

\bibitem{VGG}
Simonyan, K., Zisserman, A.: Very deep convolutional networks for large-scale
  image recognition. In: ICLR (2015)

\bibitem{Taniai}
Taniai, T., Sinha, S.N., Sato, Y.: Joint recovery of dense correspondence and
  cosegmentation in two images. In: CVPR (2016)

\bibitem{DAISY}
Tola, E., Lepetit, V., Fua, P.: Daisy: An efficient dense descriptor applied to
  wide-baseline stereo. TPAMI  (2010)

\bibitem{ufer2017deep}
Ufer, N., Ommer, B.: Deep semantic feature matching. In: CVPR (2017)

\bibitem{StereoMatch}
Wang, Z.F., Zheng, Z.G.: A region based stereo matching algorithm using
  cooperative optimization. In: CVPR (2008)

\bibitem{ObjAware}
Yang, F., Li, X., Cheng, H., Li, J., Chen, L.: Object-aware dense semantic
  correspondence. In: CVPR (2017)

\bibitem{PCK}
Yang, Y., Ramanan, D.: Articulated human detection with flexible mixtures of
  parts. TPAMI  (2013)

\bibitem{Lift}
Yi, K.M., Trulls, E., Lepetit, V., Fua, P.: Lift: Learned invariant feature
  transform. In: ECCV (2016)

\bibitem{DeepC}
Zagoruyko, S., Komodakis, N.: Learning to compare image patches via
  convolutional neural networks. In: CVPR (2015)

\bibitem{FlowWeb}
Zhou, T., Jae~Lee, Y., Yu, S.X., Efros, A.A.: Flowweb: Joint image set
  alignment by weaving consistent, pixel-wise correspondences. In: CVPR (2015)

\bibitem{3D-cycle}
Zhou, T., Krahenbuhl, P., Aubry, M., Huang, Q., Efros, A.A.: Learning dense
  correspondence via 3d-guided cycle consistency. In: CVPR (2016)

\bibitem{Multi-match}
Zhou, X., Zhu, M., Daniilidis, K.: Multi-image matching via fast alternating
  minimization. In: ICCV (2015)

\bibitem{CycleGan}
Zhu, J.Y., Park, T., Isola, P., Efros, A.A.: Unpaired image-to-image
  translation using cycle-consistent adversarial networks. In: CVPR (2017)

\bibitem{DFNet}
Zou, Y., Luo, Z., Huang, J.B.: Df-net: Unsupervised joint learning of depth and
  flow using cross-task consistency. In: ECCV (2018)

\end{thebibliography}

\end{document}